\documentclass{article}
\usepackage{pifont}
\PassOptionsToPackage{numbers,compress}{natbib}

\usepackage[preprint]{neurips_2026}

\usepackage[utf8]{inputenc} % allow utf-8 input
\usepackage[T1]{fontenc}    % use 8-bit T1 fonts
\usepackage{hyperref}       % hyperlinks
\usepackage{url}            % simple URL typesetting
\usepackage{booktabs}       % professional-quality tables
\usepackage{amsmath}        % math environments and commands
\usepackage{amsfonts}       % blackboard math symbols
\usepackage{nicefrac}       % compact symbols for 1/2, etc.
\usepackage{microtype}      % microtypography
\usepackage{xcolor}         % colors
\usepackage{graphicx}       % include graphics
\usepackage{wrapfig}        % wrap text around figures
\usepackage{afterpage}      % \afterpage{\clearpage} to flush deferred floats
\usepackage{caption}        % \captionof for captions inside minipages
\usepackage{etoc}           % local table of contents for the supplementary
\usepackage{float}          % [H] placement specifier for forced inline figures

\title{Seeing Across Skies and Streets: Feedforward 3D Reconstruction from Satellite, Drone, and Ground Images}

\newcommand{\samethanks}[1][\value{footnote}]{\footnotemark[#1]}

\author{%
  Qiwei Wang\thanks{Equal contribution.} \\
  ShanghaiTech University \\
  Shanghai, China \\
  \And
  Zhongyao Tuo\samethanks \\
  ShanghaiTech University \\
  Shanghai, China \\
  \And
  Xianghui Ze \\
  Nanjing University of Science and Technology \\
  Nanjing, China \\
  \And
  Yujiao Shi\thanks{Corresponding author.} \\
  ShanghaiTech University \\
  Shanghai, China \\
  \texttt{shiyj2@shanghaitech.edu.cn}
}

\begin{document}

\maketitle

\begin{figure}[h]
  \centering
  \includegraphics[width=\textwidth]{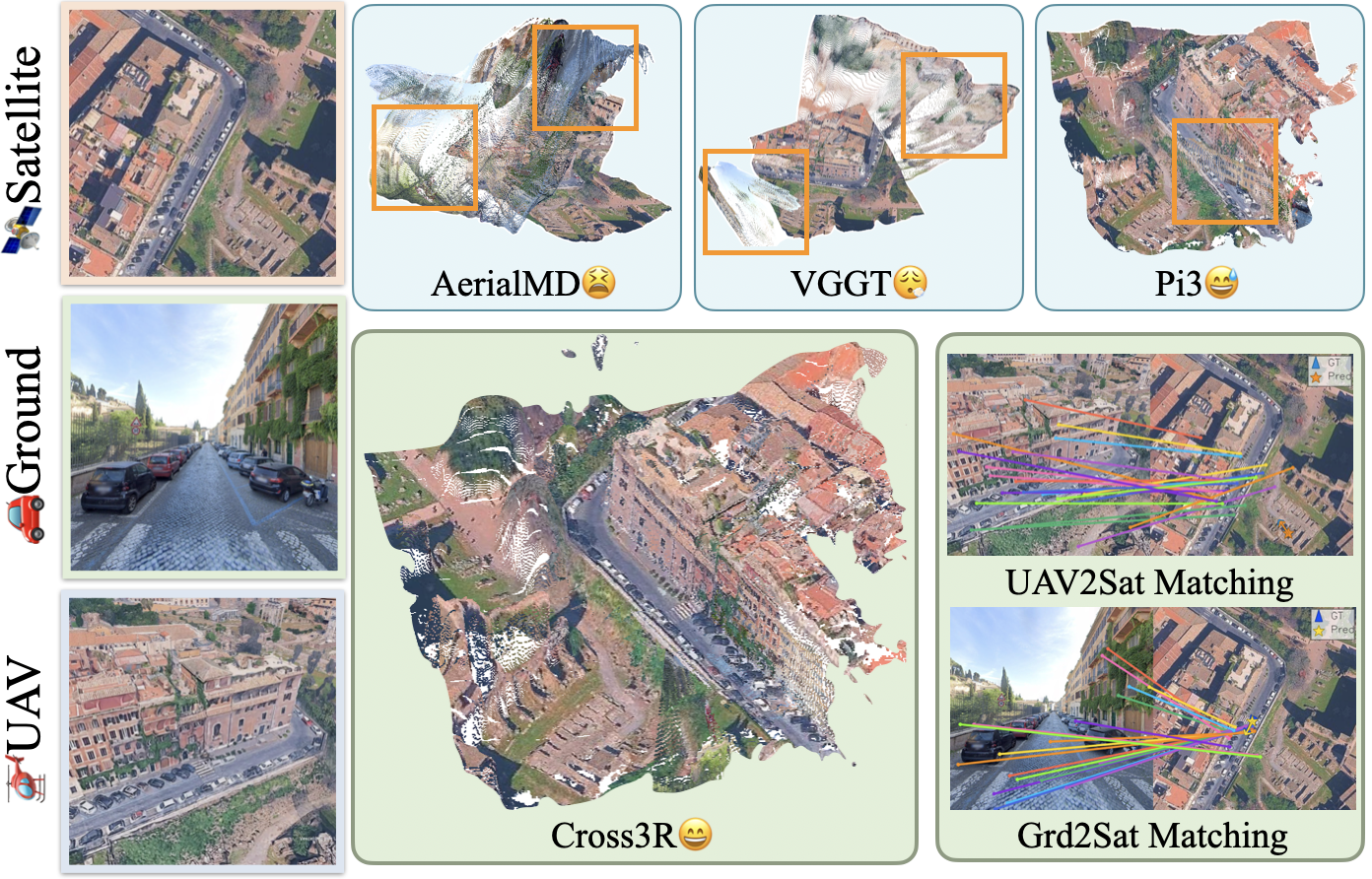}
  \caption{Cross3R ingests a satellite tile along with one or two perspective views (UAV, ground, or both) and jointly recovers a cross-view 3D point cloud, all camera poses, and the on-tile localization of each perspective camera.}
  \label{fig:teaser}
\end{figure}

\begin{abstract}
Cross-view localization classically asks: where does this ground image lie on the satellite tile? Existing methods are typically limited to 3-DoF estimates---an $(x,y)$ position and a yaw angle---because nadir satellite imagery provides no direct cues for roll, pitch, or altitude, forcing a reliance on planar-motion and zero-tilt assumptions. These assumptions break on real terrain with slopes, ramps, and tilted camera mounts. To overcome this, we introduce a single UAV image as an intermediate viewpoint: it reveals the 3D structure invisible from nadir, supplies the cues for roll, pitch, and altitude that the satellite alone cannot provide, and needs only spatial overlap with the ground camera---no known relative pose is required. Building on this insight, we propose \textbf{Cross3R}, a flexible feed-forward model that ingests a satellite tile together with a UAV image, a ground image, or both, and, in a single forward pass, recovers a cross-view 3D point cloud, the 6-DoF poses of every input camera, and the on-tile $(x,y)$ position and yaw of each perspective camera. For training and evaluation, we also construct \textbf{CrossGeo}, a 278K-image tri-view dataset spanning 85 scenes across every continent except Antarctica. On CrossGeo, Cross3R consistently outperforms feed-forward 3D baselines in point-cloud reconstruction, 6-DoF camera-pose estimation, and cross-view localization. On KITTI, it outperforms dedicated cross-view methods trained on KITTI on most metrics, despite having no KITTI training itself.
\end{abstract}

\etocdepthtag.toc{main}
\section{Introduction}
\label{sec:intro}

Ground-to-satellite camera localization aims to estimate the pose of a ground-view camera with respect to a satellite image. This paradigm is appealing because satellite imagery provides globally available, frequently updated, and cost-effective environmental coverage. In contrast, 3D maps built from LiDAR scans, structure-from-motion point clouds, meshes, or neural representations are expensive to construct and difficult to maintain, making satellite-based localization an especially scalable alternative for large-area deployments.

Despite its appeal, direct ground-to-satellite matching is challenging due to the drastic viewpoint gap, severe appearance changes, and ambiguous geometric correspondences between the two modalities. As a consequence, existing approaches~\cite{xia2023convolutional, song2023learning, shi2022accurate, lentsch2023slicematch, wang2023fine, wang2025bevsplat} restrict the problem to 3-DoF localization—recovering only the $(x, y)$ position and yaw angle of the ground camera—because a single nadir tile provides no direct cues for roll, pitch, or altitude; these methods further assume the ground camera moves on a flat plane at a fixed height with zero roll and pitch. Real environments, however, contain slopes, ramps, uneven terrain, and tilted camera mounts—conditions common in urban scenes, construction zones, and robotic deployments—where these assumptions break and 3-DoF methods degrade sharply in both accuracy and robustness. Overcoming this limitation requires constraining the missing degrees of freedom with cues that satellite imagery alone cannot supply.

Our key insight is that a single UAV image captured at an intermediate altitude naturally bridges the large viewpoint gap between the satellite (top-down) and the ground (perspective) views: it simultaneously observes the top-down layout characteristic of satellite views and the oblique 3D structure visible from the ground. This connective viewpoint is what prior 3-DoF cross-view methods could not establish from a satellite tile alone. Crucially, our approach does not require known relative poses between the UAV and the ground or satellite cameras; only a shared field of view is necessary---a condition easily satisfied by cooperative UGV--UAV systems, on-demand drones, and patrol archives indexed with coarse GPS.

Building on this insight, we propose \textbf{Cross3R}, a feed-forward cross-view reconstruction model that breaks the 3-DoF ceiling of prior cross-view methods. Cross3R accepts satellite tiles paired with a ground image, a UAV image, or both, and incorporating the UAV image yields pronounced gains in ground-to-satellite localization (Fig.~\ref{fig:uav_heatmaps}, Tab.~\ref{tab:arch_ablation}). Beyond the UAV view, two other design choices are also important. First, a per-sample altitude redefinition (Sec.~\ref{sec:training_algorithm}) stabilizes training by aligning the satellite, UAV, and ground depth ranges. Second, a single per-tile scale $\rho$ replaces satellite intrinsics. Because $\rho$ is a learned relative quantity (not true meters-per-pixel), any 3D point---including the ground- and UAV-camera centers---can be placed on the satellite tile simply by transforming to the satellite frame and dividing by $\rho$. The impact of this design is clearest on KITTI (Tab.~\ref{tab:kitti_crossarea}), where Cross3R surpasses specialized cross-view methods trained directly on the KITTI split~\cite{xia2023convolutional, shi2022accurate, lentsch2023slicematch} on most metrics. Since no public benchmark provides all three views with full 6‑DoF poses and metric depth (Tab.~\ref{tab:datasets_comparison}), we construct and release \textbf{CrossGeo}, an automatically curated dataset of 85 globally distributed scenes sourced from Google Maps, Google Earth, and Google Street View. Evaluated on CrossGeo, KITTI, and AnyVisLoc~\cite{ye2025exploring}, Cross3R substantially outperforms strong feed‑forward 3D baselines~\cite{wang2025vggt, wang2025pi, vuong2025aerialmegadepth} and dedicated 3‑DoF cross‑view methods~\cite{xia2023convolutional, shi2022accurate, lentsch2023slicematch}.

\begin{table}[!t]
\centering
\caption{Datasets Comparison}
\label{tab:datasets_comparison}
\resizebox{\textwidth}{!}{%
\begin{tabular}{lccccccccc}
\toprule
Name & Data Source & Images & Grd & UAV & Map & Grd Pose & UAV Observation  & Scenes \\
\midrule
AnyVisLoc~\cite{ye2025exploring}      & Real        & 18,000  & \ding{55} & \checkmark & \ding{55} & --    & Multi-view  & Multiple  \\
University-1652~\cite{zheng2020university} & Real  & 50,218  & \checkmark & \checkmark & \checkmark & \ding{55} & no pose     & Buildings \\
CVUSA~\cite{workman2015localize}               & Real        & 44,516  & \checkmark & \ding{55} & \checkmark & 3DoF  & --          & Streets   \\
CVACT~\cite{liu2019lending}               & Real        & 44,416  & \checkmark & \ding{55} & \checkmark & 3DoF  & --          & Streets   \\
KITTI~\cite{geiger2013vision}               & Real        & 14,999  & \checkmark & \ding{55} & \checkmark & 3DoF  & --          & Streets   \\
VIGOR~\cite{zhu2021vigor}               & Real        & 195,832 & \checkmark & \ding{55} & \checkmark & 3DoF  & --          & Multiple  \\
ULTRRA Challenge~\cite{2zs6-ht63-24}   & Synthetic   & 1,207   & \checkmark & \checkmark & \ding{55} & 6DoF  & Multi-view  & Buildings \\
AerialMD~\cite{vuong2025aerialmegadepth} & Real+Synth & 132,137 & \checkmark & \checkmark & \ding{55} & 6DoF  & Multi-view  & Multiple  \\
\textbf{Ours (CrossGeo)} & \textbf{Real+Synth} & \textbf{277,812} & \textbf{\checkmark} & \textbf{\checkmark} & \textbf{\checkmark} & \textbf{6DoF} & \textbf{Multi-view} & \textbf{Multiple} \\
\bottomrule
\end{tabular}%
}
\end{table}

\section{Method}
\label{sec:method}

We present \emph{Cross3R}, a feed-forward foundation model that jointly recovers tri-view 3D point clouds, all camera poses, and the 6-DoF cross-view localization of every input view in a single forward pass. Section~\ref{sec:dataset} describes \textbf{CrossGeo}, the large-scale satellite\,/\,UAV\,/\,ground training corpus we construct from publicly available sources. Section~\ref{sec:training_algorithm} then details how Cross3R extends $\pi^3$~\cite{wang2025pi} with three modifications targeted at the satellite branch and how the model is trained.

\subsection{CrossGeo Dataset}
\label{sec:dataset}

\begin{figure}[!t]
  \centering
  \includegraphics[width=\textwidth]{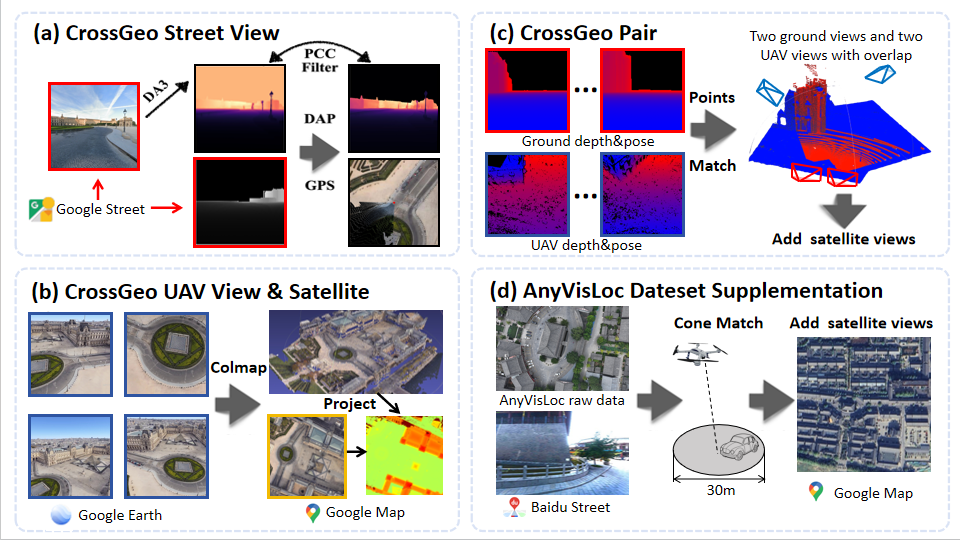}
  \caption{CrossGeo data sources. (a) Ground views and their coarse depth from \emph{Google Street View}. (b) UAV captures rendered in \emph{Google Earth} together with co-located satellite tiles from \emph{Google Maps}. (c) Tri-view pairing across the three modalities. (d) AnyVisLoc test-set augmentation.}
  \label{fig:data_sources}
  \vspace{-12pt}
\end{figure}

Training a tri-view model requires paired (satellite, UAV, ground) data with full 6-DoF poses and dense metric depth across all three views---a combination that no public benchmark currently provides (Tab.~\ref{tab:datasets_comparison}). We therefore build \textbf{CrossGeo}, a tri-view corpus of $46{,}302$ samples drawn from $85$ globally distributed scenes spanning every continent except Antarctica and covering urban, suburban, rural, and natural terrain. Each sample bundles two satellite, two UAV, and two ground views of the same region with calibrated poses and dense metric depth, so CrossGeo contains $46{,}302 \times 6 = 277{,}812$ images in total. Figure~\ref{fig:data_sources} summarizes the four data sources; Figures~\ref{fig:dataset_overview}--\ref{fig:data_pipeline} of the supplementary material visualize the resulting tri-view coverage and the construction pipeline.

\paragraph{Data collection.}
\label{sec:dataset_collection}
Acquisition is anchored on each scene's GPS center. We extract a $300\,\text{m}\!\times\!300\,\text{m}$ Google Maps satellite tile, render UAV views in Google Earth at altitudes $H_{\text{drone}}\!\in\![30, 120]$\,m and pitch angles $\in[0^{\circ}, 90^{\circ}]$, and collect ground images via the Google Street View API. The high-pitch range ($60^{\circ}$--$90^{\circ}$) is oversampled because high-altitude low-pitch renderings suffer from pixel stretching that degrades ground resolution and geometric stability; the resulting altitude/pitch distribution and the global geographic coverage are visualized in Fig.~\ref{fig:dataset_overview}(b)--(c). We use Google Street View rather than Google Earth's street-level renderings, which suffer from geometric artifacts such as texture clipping through buildings and a larger domain gap to real-world deployment.

\paragraph{Pose recovery.}
\label{sec:dataset_pose}
We define a world frame whose $z$-axis points east, $x$-axis points south, and $y$-axis points vertically downward. For UAV and ground views, the Google Earth and Street View APIs already expose every quantity needed to assemble extrinsics---altitude, GPS, yaw, pitch---which we convert directly into a rotation--translation pair $(R, T)$. Satellite poses are more involved because Google Maps exposes only the tile center's GPS. We therefore model every satellite image as a virtual camera looking straight down with a narrow $[2^{\circ}, 5^{\circ}]$ field of view, and recover its altitude by sweeping candidate values in Google Earth and stopping when the rendered top-down view at the same GPS matches the actual Google Maps tile. An altitude of $5{,}726$\,m yields near-perfect alignment and is adopted as the canonical satellite camera height.

\paragraph{Metric depth and tri-view pairing.} Each modality receives its dense metric depth through a dedicated pipeline (Fig.~\ref{fig:data_pipeline}). \emph{Ground depth} (Fig.~\ref{fig:data_sources}(a)) starts from Street View, which ships an absolute-scale but coarse depth that captures the overall scene scale yet misses fine structure (distant buildings, street lamps); we therefore run \textsc{DepthAnything\,v3}~\cite{lin2025depth} on each ground frame to obtain a sharp but \emph{relative} depth, and fuse the two through \emph{Prior Depth Anything}~\cite{wang2025depth}, which uses the Street View depth as a metric anchor for the global scale and shift of the relative prediction. As a safeguard, samples whose rescaled depth has low Pearson correlation~\cite{benesty2009pearson} with the relative prediction are discarded. \emph{UAV depth} (Fig.~\ref{fig:data_sources}(b)) is recovered by COLMAP~\cite{schonberger2016structure} multi-view stereo over a denser auxiliary capture of each scene, and \emph{satellite depth} is obtained for free by projecting the resulting point cloud into each top-down virtual camera. Given a depth and a pose for every image, we score cross-modality pairs by voxel overlap and take the highest-scoring six-image tuples (two views per modality) as our tri-view samples (Fig.~\ref{fig:data_sources}(c)), requiring all $\binom{6}{2}{=}15$ pairwise overlaps to be non-empty.

\paragraph{Train / val / test split.}
\label{sec:dataset_splits}
The 85 scenes are partitioned at the \emph{scene} level into three disjoint subsets, ensuring that no city appears in more than one split. The training split comprises $75$ scenes ($38{,}962$ samples); the validation split contains $5$ scenes ($3{,}614$ samples) drawn from North American cities; and the test split consists of $5$ scenes ($3{,}726$ samples) sourced from cities with no overlap with the training set---probing generalization across both city and continent boundaries. This cross-continent test split serves as the in-distribution evaluation set used throughout Section~\ref{sec:experiments}.

\paragraph{Real-world OOD test set.} CrossGeo's UAV imagery is rendered in Google Earth and is therefore synthetic. To stress-test Cross3R under real aerial photography, we repurpose AnyVisLoc~\cite{ye2025exploring}, a benchmark of UAV photographs captured by physical drones, as an out-of-distribution test set (Fig.~\ref{fig:data_sources}(d)). AnyVisLoc provides only the UAV view, so we augment every sample with the two missing modalities: for each AnyVisLoc UAV photograph we draw a Baidu Maps ground image within $30$\,m of the UAV's GPS, then center a $110\,\text{m}\!\times\!110\,\text{m}$ Google Maps tile on the ground camera with a random $\pm 20$\,m east--south shift to mimic deployment-time GPS uncertainty. The resulting $5{,}068$ tri-view samples form a fully real-world test set used throughout the experiments.

\begin{figure}[t]
  \centering
  \includegraphics[width=\textwidth]{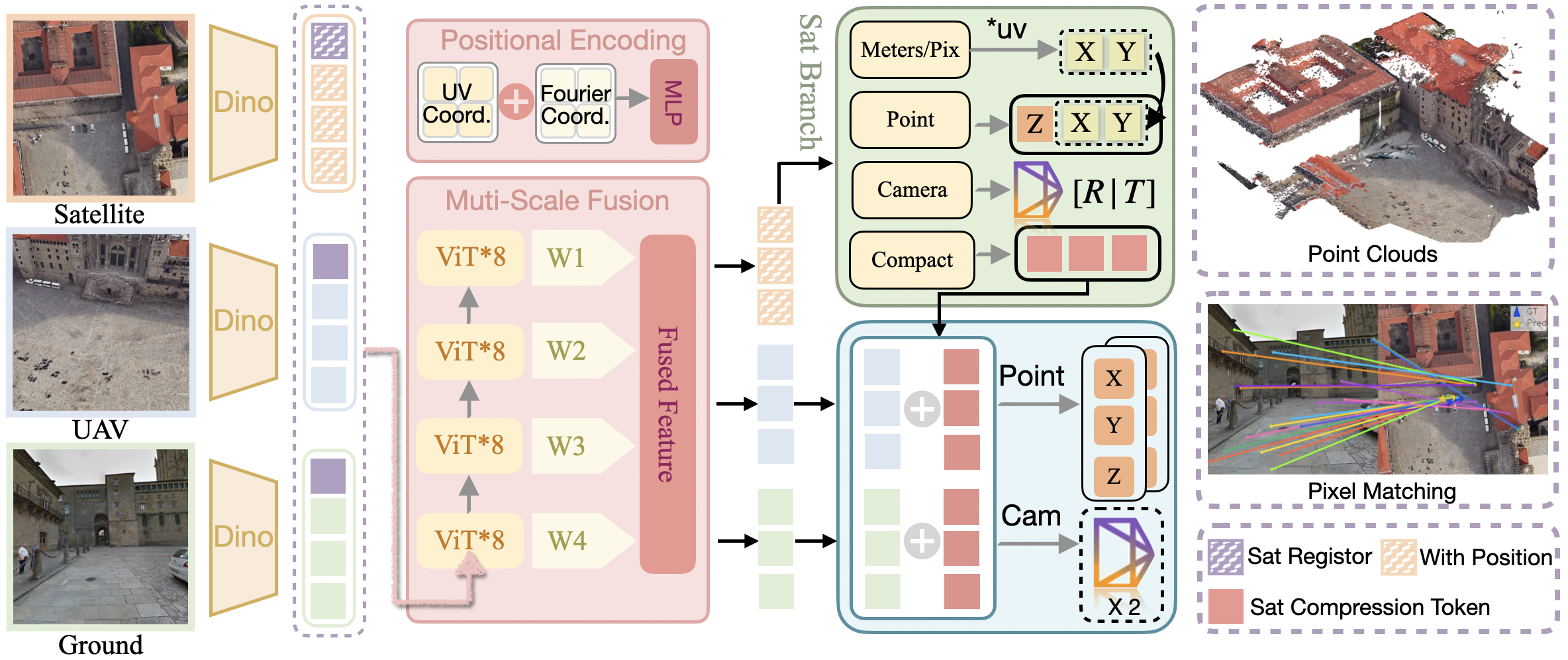}
  \caption{Overview of Cross3R. Satellite, UAV, and ground images are encoded and processed through an Alternating Attention module. Task-specific heads then predict: an orthographic representation (a per-tile scale $\rho$ and per-pixel depth $z$) for the satellite view, local 3D point maps $(x, y, z)$ for the UAV and ground views, and unified camera poses $(R, T)$ for all views.}
  \label{fig:framework}
\end{figure}

\subsection{The Cross3R Model}
\label{sec:training_algorithm}

Cross3R takes a satellite tile $I_s$ and up to two perspective views (a UAV image $I_u$ and/or a ground image $I_g$) that share overlapping scene content. In a single forward pass it predicts, for every input view $v$, a per-pixel point map $P^v \in \mathbb{R}^{H \times W \times 3}$ and a 6-DoF camera pose $(R^v, T^v)$, along with a per-image scalar $\rho$ that ties the per-pixel scale of the satellite tile to the rest of the scene. Since $\rho$ is a learned relative scale (not true meters-per-pixel), Cross3R recovers a self-consistent scene; any 3D point---including the ground- and UAV-camera centers---can then be located on the satellite tile simply by transforming to the satellite's local frame and dividing the $(x,y)$ coordinate by $\rho$. In contrast, existing feed-forward 3D models~\cite{wang2025pi, wang2025vggt, vuong2025aerialmegadepth} are designed for image collections captured from roughly similar altitudes. On satellite--UAV--ground triplets, two difficulties surface (Tab.~\ref{tab:sota_localization}, Fig.~\ref{fig:teaser}): the predicted point clouds are individually plausible but drift apart and fail to align; moreover, these models output only point clouds and extrinsics, so projecting a ground or UAV camera onto the satellite tile requires inferring satellite intrinsics from an extremely narrow field of view---a fragile step that often returns wrong intrinsics and breaks cross-view localization.

Cross3R realizes $\rho$ and the joint registration described above by building on $\pi^3$~\cite{wang2025pi} and introducing three coordinated modifications to the encoder, the alternating-attention stack, and the decoding heads (Fig.~\ref{fig:framework}).
\emph{Encoder.} We inject a Fourier positional embedding~\cite{mildenhall2021nerf} of the satellite's normalized pixel grid directly into the satellite tokens, so the orthographic grid structure enters the network from the very first layer:
\begin{equation}
    \mathrm{PE}(x) \;=\; W\!\left[\, x \,\Vert\, \sin(2\pi B^{\top} x) \,\Vert\, \cos(2\pi B^{\top} x) \,\right] + b,
\label{eq:fourier-pe}
\end{equation}
where $x \in [0,1]^2$ is a normalized pixel coordinate, $B \in \mathbb{R}^{2 \times K}$ is a fixed Gaussian random projection controlling the frequency bandwidth, and $W, b$ are learnable parameters. We also double the size of $\pi^3$'s register-token bank, giving the encoder extra capacity to disentangle satellite from perspective tokens.
\emph{Alternating attention.} Instead of attending only to the last layer, the heads fuse hidden states across layers with softmax-weighted gates, gaining simultaneous access to fine local geometry (early layers) and coarse global layout (late layers).
\emph{Decoding.} The satellite and perspective branches diverge. For the satellite branch we enforce an explicit orthographic geometric prior: a Google Maps tile is rendered by a virtual pinhole camera looking straight down with a very narrow $[2^{\circ}, 5^{\circ}]$ field of view from $5{,}726$\,m altitude (Section~\ref{sec:dataset_pose}). Although the tile is fundamentally a perspective rendering, the parallax across a single tile is sub-pixel, making an orthographic projection an excellent approximation. Under this model, every 3D point in the tile's local frame satisfies
\begin{equation}
    (x_i, y_i) \;=\; (u_i, v_i) \cdot \rho,
\label{eq:ortho-constraint}
\end{equation}
where $(u_i, v_i)$ is a pixel coordinate and $\rho$ is the aforementioned learned relative scale. A tiny MLP regresses $\rho$ per satellite image, and we replace the $(x, y)$ output of the shared point head with $(u, v) \cdot \rho$, retaining only the per-pixel depth $\hat{z}$. The resulting satellite point cloud is internally scale-consistent by construction and shares the same relative scale as the rest of the scene. For the UAV and ground branches, satellite features from the same stage are compressed by a small MLP and broadcast-added into the perspective tokens before the shared point and camera heads, giving each perspective branch explicit guidance from the satellite coordinate frame. Without such an anchor, earlier methods must resort to fragile intrinsic estimation---precisely the failure mode we observe in the baselines.

\paragraph{Loss functions.} Following $\pi^3$~\cite{wang2025pi}, the training objective is a single weighted sum of four terms,
\begin{equation}
    \mathcal{L} \;=\; \mathcal{L}_{\text{geo}} + \lambda_n \mathcal{L}_{\text{norm}} + \lambda_c \mathcal{L}_{\text{conf}} + \lambda_p \mathcal{L}_{\text{cam}}.
\end{equation}
$\mathcal{L}_{\text{geo}}$ first solves for an optimal global scale $s^\star$ that aligns predicted to ground-truth point maps and then penalizes the depth-weighted $L_1$ residual at that scale; $\mathcal{L}_{\text{norm}}$ minimizes the angular error between predicted and ground-truth surface normals to encourage locally smooth geometry; $\mathcal{L}_{\text{conf}}$ supervises a per-pixel confidence map with binary cross-entropy, with target $1$ when the scaled reconstruction error falls below a threshold $\epsilon$; and $\mathcal{L}_{\text{cam}}$ supervises relative poses between all ordered view pairs with a geodesic rotation loss and a Huber translation loss on scale-corrected predictions. We set $\lambda_n\!=\!1$, $\lambda_c\!=\!0.05$, $\lambda_p\!=\!0.1$, and enable $\mathcal{L}_{\text{norm}}$ only after a five-epoch warm-up.

\paragraph{Per-sample altitude redefinition.}
\begin{wrapfigure}{r}{0.5\textwidth}
  % \vspace{-10pt}
  \centering
  \includegraphics[width=0.5\textwidth]{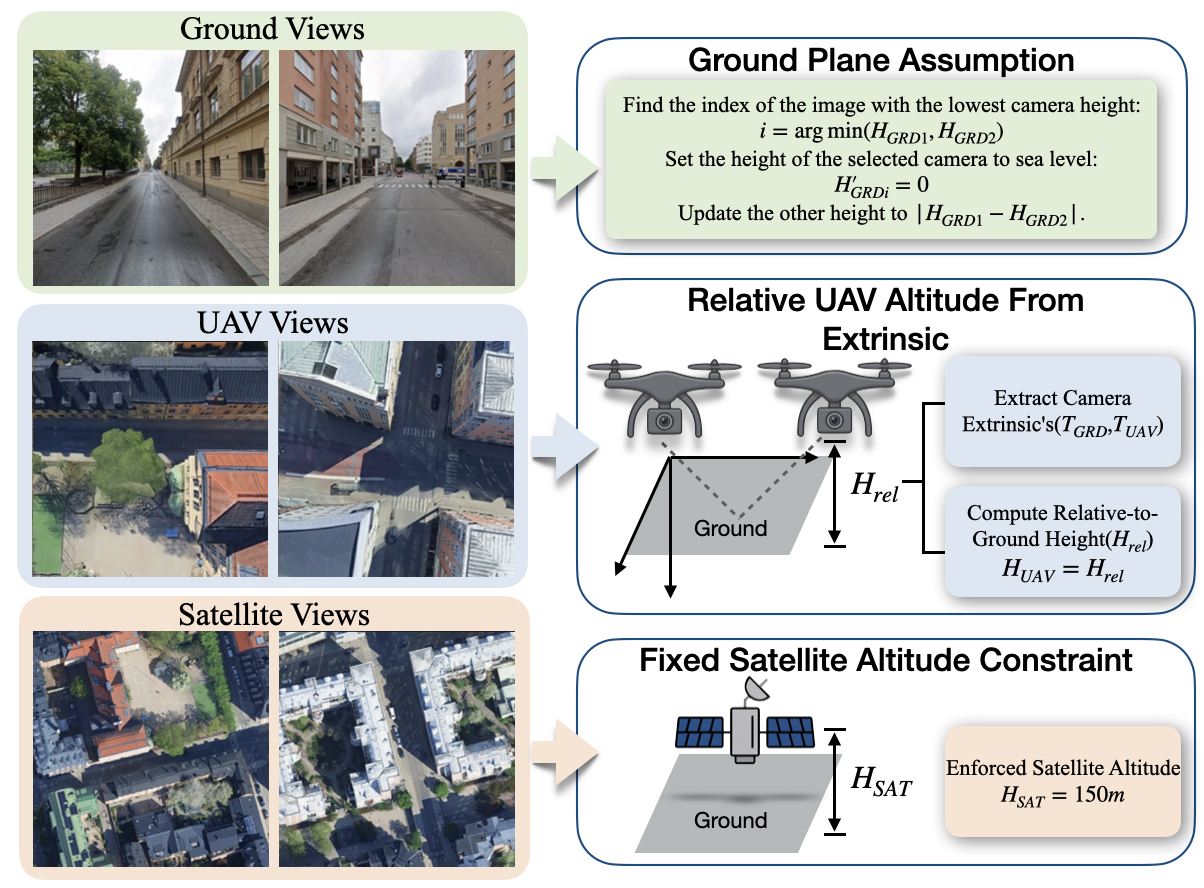}
  \caption{Per-sample altitude redefinition.}
  \label{fig:training_setting}
  \vspace{-10pt}
\end{wrapfigure}
Raw satellite poses in CrossGeo are anchored at $5{,}726$\,m above ground (Section~\ref{sec:dataset_pose}), and back-propagating through translations that large destabilizes training. We therefore re-anchor altitudes locally at dataset-preparation time. CrossGeo is collected as image pairs, each containing two ground, two UAV, and two satellite views of overlapping scene content; within a pair we take the lower of the two ground cameras as the reference and place it at $y\!=\!0$. The second ground camera then sits at its relative altitude $H_{\text{grd}}^{\prime}$ above this reference (a small offset on uneven terrain), the two UAV cameras at their relative altitudes $H_{\text{drone}}\!\in\![30, 120]$\,m, and the two satellite cameras at a constant $H_{\text{sat}}\!=\!150$\,m, chosen just above the maximum UAV flight altitude (Figure~\ref{fig:training_setting}). This keeps gradient magnitudes through the satellite translation bounded and gives the network a per-sample-constant satellite-to-ground depth prior that, in our experience, accelerates convergence.

\section{Experiments}
\label{sec:experiments}

We benchmark Cross3R on two complementary task families. Section~\ref{sec:exp_tri} evaluates \emph{cross-altitude point cloud reconstruction and camera pose estimation}, where we recover dense geometry and the relative camera poses jointly across the satellite, UAV, and ground views. Section~\ref{sec:exp_loc} then evaluates \emph{cross-view ground-camera localization and pixel matching}, measuring how accurately the ground camera is placed on the satellite tile and how reliably pixel-level correspondences can be established across the three modalities. Section~\ref{sec:ablations} additionally presents three ablation studies: one examining how the UAV pose---altitude, pitch, and horizontal distance to the ground camera---affects the UAV's contribution to ground-camera localization, another isolating the contribution of each architectural component, and a third measuring how Cross3R scales with the amount of CrossGeo training data.

\paragraph{Setup.} We initialize Cross3R from the released $\pi^3$~\cite{wang2025pi} checkpoint, inheriting all weights of layers shared with $\pi^3$, and fine-tune the model on CrossGeo (Section~\ref{sec:dataset}). Despite the architectural changes described in Section~\ref{sec:training_algorithm}, our additions are deliberately lightweight: Cross3R adds only ${\sim}\,2.5$\,M parameters on top of $\pi^3$ ($961.20$\,M vs.\ $958.70$\,M, a $0.26\%$ increase), and per-triplet inference time is essentially unchanged ($137.8$\,ms vs.\ $\pi^3$'s $138.5$\,ms). Following $\pi^3$, fine-tuning runs in two stages on $8\times$ NVIDIA L40 GPUs: stage 1 uses $224\times 224$ inputs for 6 hours with a per-GPU batch size of 64, and stage 2 uses variable resolutions in $[100{,}000, 255{,}000]$ pixels with aspect ratios drawn from $[0.5, 2.0]$ for 8 hours with a per-GPU batch size of 16. Each batch dynamically groups 2--3 frames; at inference, all images are resized to $504\times 504$.

% Force the 4 tables to defer to the next page (page 7), so page 6 stays text-only.
\suppressfloats[t]

% Tight spacing for the 4-table block at the top of the experiments page.
\begingroup
\setlength{\textfloatsep}{-12pt plus 0.5pt minus 0.5pt}
\setlength{\floatsep}{-18pt plus 0.5pt minus 0.5pt}
\setlength{\intextsep}{-6pt plus 0.5pt minus 0.5pt}
\setlength{\abovecaptionskip}{-4pt}
\setlength{\belowcaptionskip}{-10pt}
\renewcommand{\arraystretch}{0.78}
\captionsetup{font=small,skip=0pt,belowskip=-6pt}

\begin{table}[t]
\centering
\caption{Point cloud reconstruction and camera pose estimation on CrossGeo. $\pi^{3*}$ denotes $\pi^3$ fine-tuned on CrossGeo. Best in \textbf{bold}, second-best \underline{underlined}.}
\label{tab:sota_pcd_pose}
\resizebox{\textwidth}{!}{%
\begin{tabular}{lc|ccc|ccc|ccc|c}
\toprule
 & Acc & \multicolumn{3}{c|}{$\delta$ (\%)} & \multicolumn{3}{c|}{RRA (\%)} & \multicolumn{3}{c|}{RTA (\%)} & AUC \\
\cmidrule(lr){3-5} \cmidrule(lr){6-8} \cmidrule(lr){9-11}
Method & mean$\downarrow$ & $0.5$m & $1$m & $2$m & $5^{\circ}$ & $15^{\circ}$ & $25^{\circ}$ & $5^{\circ}$ & $15^{\circ}$ & $25^{\circ}$ & $30^{\circ}$ \\
\midrule
AerialMD~\cite{vuong2025aerialmegadepth} & 2.1334 & 28.48 & 48.56 & 70.12 & 3.00 & 47.00 & 62.50 & 7.50 & 54.50 & 70.00 & 35.41 \\
VGGT~\cite{wang2025vggt}                & 2.3029 & 18.59 & 33.97 & 57.45 & 8.40 & 47.75 & 57.76 & 0.00 & 1.61 & 8.30 & 2.69 \\
$\pi^3$~\cite{wang2025pi}                   & 1.9818 & 24.86 & 42.31 & 65.47 & 26.45 & 64.51 & 72.97 & 1.34 & 16.43 & 32.44 & 15.98 \\
$\pi^{3*}$                                & \underline{1.1771} & \underline{39.18} & \underline{61.40} & \underline{83.55} & \underline{86.99} & \underline{95.40} & \underline{96.15} & \underline{29.93} & \underline{93.28} & \underline{98.98} & \underline{71.87} \\
\textbf{Cross3R (Ours)}                  & \textbf{1.0671} & \textbf{43.52} & \textbf{66.56} & \textbf{86.23} & \textbf{95.10} & \textbf{97.91} & \textbf{98.23} & \textbf{41.19} & \textbf{95.66} & \textbf{99.52} & \textbf{76.84} \\
\bottomrule
\end{tabular}}
\vspace{-12pt}
\end{table}

\begin{table}[t]
\centering
\caption{Cross-view ground- and UAV-camera localization on CrossGeo. Meter (m) is the on-tile translation error of the predicted camera position from ground truth; Yaw ($^{\circ}$) is the rotation error around the up axis.}
\label{tab:sota_localization}
\resizebox{\textwidth}{!}{%
\begin{tabular}{lcccccc|cccccc}
\toprule
 & \multicolumn{6}{c|}{Ground Camera} & \multicolumn{6}{c}{UAV Camera} \\
\cmidrule(lr){2-7} \cmidrule(lr){8-13}
Method & \shortstack{Meter\\Mean$\downarrow$} & \shortstack{Meter\\Med$\downarrow$} & \shortstack{Yaw\\Mean$\downarrow$} & \shortstack{Yaw\\Med$\downarrow$} & \shortstack{PCK\\@2m$\uparrow$} & \shortstack{PCK\\@5m$\uparrow$} & \shortstack{Meter\\Mean$\downarrow$} & \shortstack{Meter\\Med$\downarrow$} & \shortstack{Yaw\\Mean$\downarrow$} & \shortstack{Yaw\\Med$\downarrow$} & \shortstack{PCK\\@2m$\uparrow$} & \shortstack{PCK\\@5m$\uparrow$} \\
\midrule
AerialMD~\cite{vuong2025aerialmegadepth} & 22.22 & 14.70 & 36.19 & 11.38 & 4.92 & 18.54 & 12.77 & 10.69 & 10.06 & 4.14 & 25.88 & \underline{66.77} \\
VGGT~\cite{wang2025vggt}                & 121.42 & 117.02 & 37.48 & 20.89 & 0.02 & 0.09 & 12.12 & \underline{6.81} & 11.18 & 4.78 & \underline{29.19} & 63.84 \\
$\pi^3$~\cite{wang2025pi}                   & 62.84 & 60.96 & 25.84 & 8.12 & 0.96 & 3.96 & 14.82 & 7.86 & 9.81 & 4.24 & 11.26 & 34.93 \\
$\pi^{3*}$                                & \underline{7.16} & \underline{5.19} & \underline{6.51} & \underline{2.05} & \underline{11.04} & \underline{42.62} & \underline{12.06} & 8.77 & \underline{2.99} & \underline{1.42} & 8.66 & 33.33 \\
\textbf{Cross3R (Ours)}                  & \textbf{3.68} & \textbf{2.14} & \textbf{3.66} & \textbf{1.30} & \textbf{42.60} & \textbf{76.34} & \textbf{2.38} & \textbf{1.64} & \textbf{1.92} & \textbf{0.87} & \textbf{67.75} & \textbf{91.23} \\
\bottomrule
\end{tabular}}
\vspace{-12pt}
\end{table}

\begin{table}[t]
\centering
\caption{Zero-shot cross-view ground-camera localization on KITTI Test\,2 ($n{=}7{,}542$, unknown orientation; inputs: ground and satellite images only). Upper block: 2D-only baselines from CCVPE~\cite{xia2023convolutional}; lower block: multi-view 3D reconstruction models. Best in \textbf{bold}, second-best \underline{underlined}.}
\label{tab:kitti_crossarea}
\resizebox{\textwidth}{!}{%
\begin{tabular}{lcc|cc|ccc|ccc|cc}
\toprule
& \multicolumn{2}{c|}{Loc.\,(m)$\downarrow$} & \multicolumn{2}{c|}{Yaw\,(deg)$\downarrow$} & \multicolumn{3}{c|}{Lat.\,recall\,(\%)$\uparrow$} & \multicolumn{3}{c|}{Lon.\,recall\,(\%)$\uparrow$} & \multicolumn{2}{c}{Ori.\,recall\,(\%)$\uparrow$} \\
\cmidrule(lr){2-3} \cmidrule(lr){4-5} \cmidrule(lr){6-8} \cmidrule(lr){9-11} \cmidrule(lr){12-13}
Method & Mean & Median & Mean & Median & @1m & @3m & @5m & @1m & @3m & @5m & @1$^{\circ}$ & @3$^{\circ}$ \\
\midrule
\multicolumn{13}{l}{\textit{Traditional cross-view localization\,---\,2D position only}} \\
LM~\cite{shi2022accurate}                & 15.50 & 16.02 & 89.84 & 89.85 & 5.60 & 16.02 & 25.60 & 5.64 & 15.86 & 25.76 & 0.60 & 1.60 \\
SliceMatch~\cite{lentsch2023slicematch}  & 14.85 & 11.85 & \underline{23.64} & 7.96 & \underline{24.00} & \underline{62.52} & \underline{72.89} & \underline{7.17} & \underline{26.11} & \underline{33.12} & \textbf{31.69} & \underline{31.69} \\
CCVPE~\cite{xia2023convolutional}        & \underline{13.94} & \underline{10.98} & 77.84 & 63.84 & 23.42 & 49.15 & 60.46 & \textbf{11.81} & \textbf{29.85} & \textbf{42.12} & 4.39 & 14.09 \\
\midrule
\multicolumn{13}{l}{\textit{Multi-view 3D reconstruction\,---\,camera pose $+$ dense point cloud}} \\
$\pi^3$~\cite{wang2025pi}                    & 48.26 & 42.81 & 27.67 & 13.45 & 4.18 & 10.69 & 18.10 & 2.85 & 7.45 & 11.92 & 5.56 & 16.31 \\
$\pi^{3*}$                                 & 16.28 & 14.85 & \textbf{23.16} & \underline{4.09} & 9.44 & 27.61 & 44.18 & 4.34 & 12.86 & 21.36 & 15.27 & \underline{40.27} \\
\textbf{Cross3R (Ours)}                   & \textbf{11.69} & \textbf{10.52} & 24.27 & \textbf{3.51} & \textbf{39.11} & \textbf{76.32} & \textbf{85.95} & 3.50 & 10.70 & 19.07 & \underline{16.14} & \textbf{44.90} \\
\bottomrule
\end{tabular}}
\vspace{-12pt}
\end{table}

\begin{table}[t]
\centering
\caption{Out-of-distribution cross-view ground- and UAV-camera localization on AnyVisLoc~\cite{ye2025exploring} ($n{=}5{,}068$; inputs: satellite, UAV, and ground images). Best in \textbf{bold}, second-best \underline{underlined}.}
\label{tab:sota_anyvisloc}
\resizebox{\textwidth}{!}{%
\begin{tabular}{lcccc|cccc}
\toprule
& \multicolumn{4}{c|}{Ground Camera} & \multicolumn{4}{c}{UAV Camera} \\
\cmidrule(lr){2-5} \cmidrule(lr){6-9}
Method & Meter-Mean$\downarrow$ & Meter-Med$\downarrow$ & Yaw-Mean$\downarrow$ & Yaw-Med$\downarrow$ & Meter-Mean$\downarrow$ & Meter-Med$\downarrow$ & Yaw-Mean$\downarrow$ & Yaw-Med$\downarrow$ \\
\midrule
AerialMD~\cite{vuong2025aerialmegadepth} & 48.90  & 43.23  & 59.81           & 42.56          & 40.17           & 23.25          & 71.62           & 69.40 \\
VGGT~\cite{wang2025vggt}                & 100.51 & 97.57  & 30.61           & 17.33          & 71.77           & 56.39          & 44.25           & 26.54 \\
$\pi^3$~\cite{wang2025pi}                   & 57.36  & 51.80  & 15.76           & 7.00           & 58.31           & 51.82          & 59.20           & 28.56 \\
$\pi^{3*}$                                & \underline{10.53} & \underline{8.36} & \underline{13.13} & \underline{3.92} & \underline{16.49} & \underline{14.19} & \textbf{38.11} & \textbf{11.24} \\
\textbf{Cross3R (Ours)}                  & \textbf{10.49} & \textbf{8.10} & \textbf{8.00} & \textbf{3.25} & \textbf{14.51} & \textbf{12.98} & \underline{39.39} & \underline{12.27} \\
\bottomrule
\end{tabular}}
\end{table}
\endgroup

\subsection{Cross-Altitude Point Cloud Reconstruction and Camera Pose Estimation}
\label{sec:exp_tri}

We first evaluate Cross3R on the CrossGeo test set (3,726 tri-view triplets, $504\times504$ inputs) for point-cloud quality and relative camera pose (Tab.~\ref{tab:sota_pcd_pose}). Following AerialMD~\cite{vuong2025aerialmegadepth}, point-cloud quality is assessed with Acc-mean~\cite{jensen2014large} ($\downarrow$, mean nearest-neighbor distance to ground truth) and $\delta@\tau$~\cite{eigen2014depth} ($\uparrow$, fraction of points within $\tau\in\{0.5,1,2\}$\,m); camera pose quality uses RRA and RTA~\cite{jin2021image} at $\theta\in\{5^{\circ},15^{\circ},25^{\circ}\}$ (relative rotation/translation-direction recall under $\theta$) and AUC@$30$~\cite{yi2018learning} ($\uparrow$, normalized area under the cumulative pose-accuracy curve from $0^{\circ}$ to $30^{\circ}$). Baselines include the recent feed‑forward models VGGT~\cite{wang2025vggt} and $\pi^3$~\cite{wang2025pi}, together with AerialMD~\cite{vuong2025aerialmegadepth}---a UAV/ground-specialized model obtained by fine-tuning Dust3R~\cite{wang2024dust3r}. Because Cross3R builds on $\pi^3$, we additionally fine-tune $\pi^3$ on CrossGeo under the same schedule (denoted $\pi^{3*}$), such that $\pi^3$\,$\to$\,$\pi^{3*}$ captures the contribution of our training data while $\pi^{3*}$\,$\to$\,Cross3R isolates the effect of our architecture. Both $\pi^{3*}$ and Cross3R are trained with our redefined satellite altitude instead of the raw map-tile altitude; as Tab.~\ref{tab:arch_ablation} shows, training with the raw altitude is unstable and led to divergence in our runs, so it is not a viable setting, and $\pi^{3*}$ inherits the same redefinition. All numbers are obtained from a single forward pass per triplet, with no per-scene optimization.

Cross3R leads every column of Tab.~\ref{tab:sota_pcd_pose}. $\pi^3$\,$\to$\,$\pi^{3*}$ alone delivers a large gain in both Acc-mean and AUC@$30$, attributing the bulk of the improvement to the CrossGeo data; $\pi^{3*}$\,$\to$\,Cross3R then adds substantial further gains on RRA@$5^{\circ}$ and RTA@$5^{\circ}$, attributing the remainder to the satellite-aware architecture. These same predicted point clouds and poses are reused, without any further processing, for the cross-view localization and pixel-matching evaluations in the next subsection.

\subsection{Cross-View Camera Localization and Pixel Matching}
\label{sec:exp_loc}

\begingroup
\setlength{\textfloatsep}{-4pt plus 0.5pt minus 0.5pt}
\setlength{\floatsep}{-4pt plus 0.5pt minus 0.5pt}
\setlength{\intextsep}{-4pt plus 0.5pt minus 0.5pt}
\setlength{\abovecaptionskip}{0pt}
\setlength{\belowcaptionskip}{-2pt}
\begin{figure}[t]
  \centering
  \vspace{-4pt}
  \includegraphics[width=\textwidth]{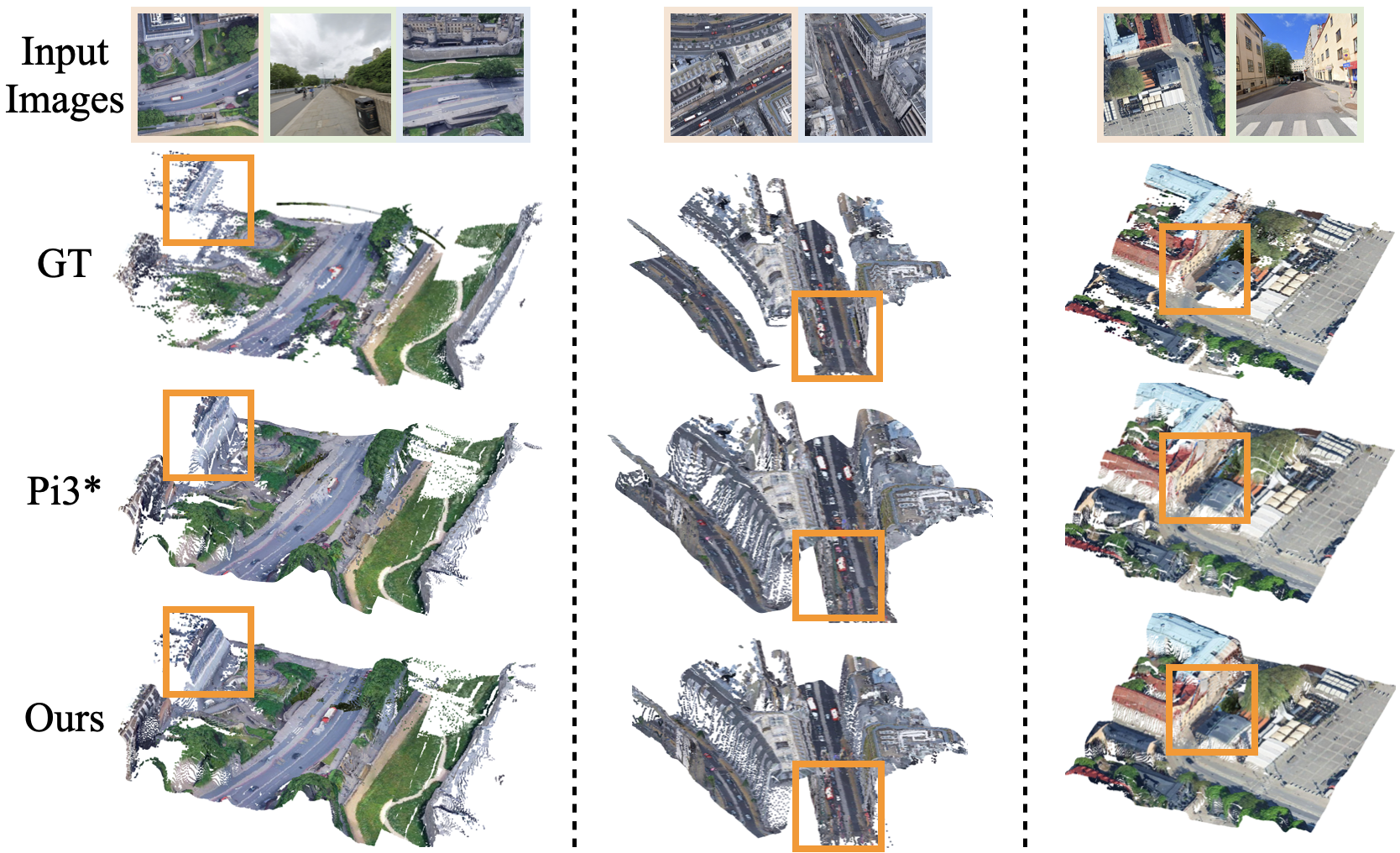}
  \caption{Predicted point clouds on three CrossGeo samples (left to right: satellite$+$ground$+$UAV, satellite$+$UAV, satellite$+$ground). Compared to $\pi^3$, Cross3R produces fewer holes and less misalignment in the regions highlighted by the orange boxes.}
  \label{fig:qualitative_points}
  \vspace{-5pt}
\end{figure}

\begin{figure}[t]
  \centering
  \vspace{-4pt}
  \includegraphics[width=\textwidth]{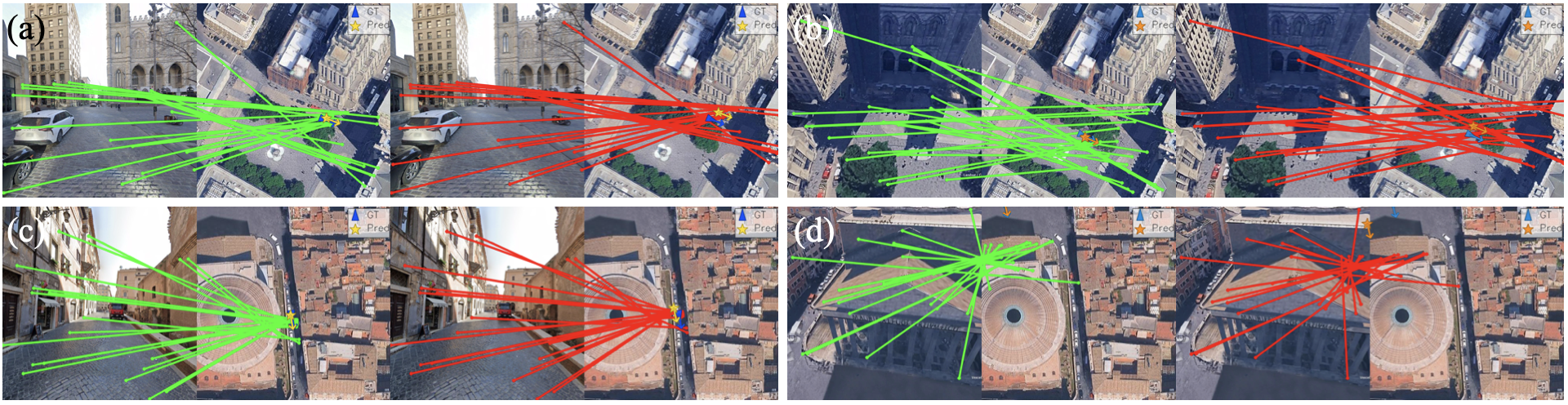}
  \caption{Cross-view localization and pixel matching on four CrossGeo samples (left two: ground--satellite; right two: UAV--satellite). Within each sample, the left column is Cross3R and the right is $\pi^{3*}$; green / red lines mark successful / failed matches. Cross3R produces more accurate results.}
  \label{fig:qualitative_matching}
  \vspace{-12pt}
\end{figure}
\endgroup
\vspace{-8pt}

We next evaluate cross-view localization---placing the ground camera on the satellite tile---and pixel-level matching across the three modalities. Cross3R produces these estimates directly from the predicted scale $\rho$ and per-view poses, with no satellite intrinsics. We probe three datasets: in-distribution CrossGeo ($3{,}726$ triplets), out-of-distribution AnyVisLoc~\cite{ye2025exploring} ($5{,}068$ triplets, with the Google Maps and Baidu Maps modalities we add in Section~\ref{sec:dataset_splits}), and zero-shot KITTI Test\,2~\cite{shi2022accurate} ($n{=}7{,}542$, no UAV view). For the KITTI dataset, following the cross-view benchmark of~\cite{shi2022accurate}, we use the left-stereo image ($90^{\circ}$ horizontal FoV) paired with Google Maps tiles at $\sim\!0.20$\,m/px that each cover roughly $100\,\text{m}\!\times\!100\,\text{m}$, with the ground camera assumed to lie within a $40\,\text{m}\!\times\!40\,\text{m}$ area at the tile center. Test\,2 contains regions geographically disjoint from the training split, and we evaluate under the harder \emph{no-orientation-prior} setting, where each aerial tile is rotated by a random angle drawn from the full $360^{\circ}$ circular domain. On every dataset we report Meter-Mean/Med (m) and Yaw-Mean/Med ($^{\circ}$); on CrossGeo we additionally report PCK@$\tau$~\cite{yang2011articulated} for $\tau\!\in\!\{2,5\}$\,m, which doubles as a pixel-matching score; on KITTI Test\,2 we follow CCVPE~\cite{xia2023convolutional} with lateral/longitudinal recall at $1$/$3$/$5$\,m and orientation recall at $1^{\circ}$/$3^{\circ}$. Beyond the multi-view 3D baselines from Sec.~\ref{sec:exp_tri}, we include three orientation-prior-free cross-view methods---LM~\cite{shi2022accurate}, SliceMatch~\cite{lentsch2023slicematch}, and CCVPE~\cite{xia2023convolutional}---trained directly on the KITTI cross-view split; their numbers are taken from the CCVPE paper.

On CrossGeo (Tab.~\ref{tab:sota_localization}) Cross3R leads all twelve columns: it halves $\pi^{3*}$'s ground translation error, nearly quadruples ground PCK@$2$m, and substantially shrinks the UAV translation error. On KITTI Test\,2 (Tab.~\ref{tab:kitti_crossarea})---trained only on CrossGeo, never on KITTI---Cross3R still attains the lowest median translation error and the strongest lateral recall at every threshold, out-performing LM, SliceMatch, and CCVPE on most metrics despite their direct supervision on KITTI. On real-world AnyVisLoc (Tab.~\ref{tab:sota_anyvisloc}) Cross3R again has the best ground and UAV translation and the best ground yaw, with only UAV yaw trailing $\pi^{3*}$ by a small margin. Fig.~\ref{fig:qualitative_matching} shows geometrically consistent satellite--UAV--ground pixel correspondences in the wild and Fig.~\ref{fig:qualitative_points} the corresponding tri-view point clouds.

\subsection{Ablation Studies}
\label{sec:ablations}

\paragraph{When does the UAV help ground-camera localization?} For every test sample, we run Cross3R twice—once with the full satellite–UAV–ground triplet and once with only satellite and ground—and measure the reduction in ground-camera error when the UAV is present (Figure~\ref{fig:uav_heatmaps}). On average, including the UAV lowers both the median translation error and the median yaw error; this is also corroborated by Tab.~\ref{tab:arch_ablation}. The benefit depends on geometry: the UAV helps most when it flies at a moderate altitude, adopts a near-nadir viewpoint, and remains within a short horizontal distance of the ground camera. This contribution directly verifies our central insight—the UAV acts as a geometric bridge that supplies the vertical structure missing from satellite-only views, so cross-view ground-camera localization works reliably on a much wider range of scenes than before.

\begingroup
\setlength{\textfloatsep}{-4pt plus 0.5pt minus 0.5pt}
\setlength{\floatsep}{0pt plus 0.5pt minus 0.5pt}
\setlength{\intextsep}{-4pt plus 0.5pt minus 0.5pt}
\setlength{\abovecaptionskip}{0pt}
\setlength{\belowcaptionskip}{-2pt}
\begin{figure}[!t]
  \centering
  \vspace{-4pt}
  \includegraphics[width=0.49\textwidth]{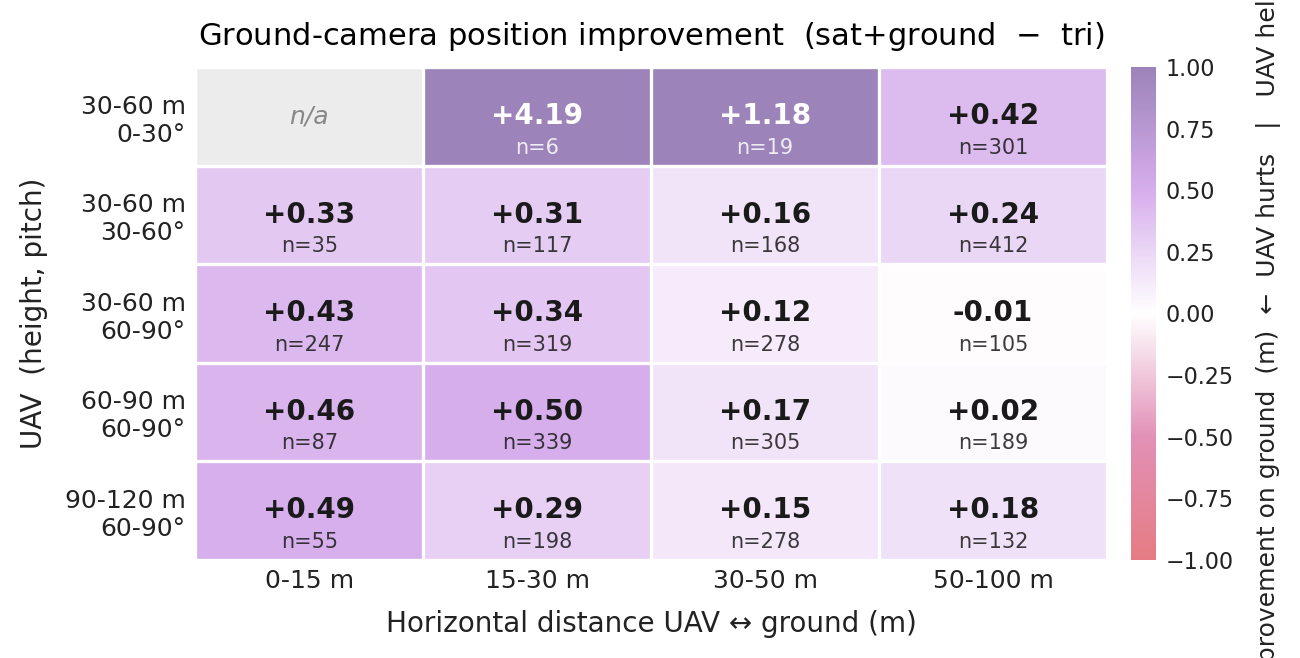}\hfill
  \includegraphics[width=0.49\textwidth]{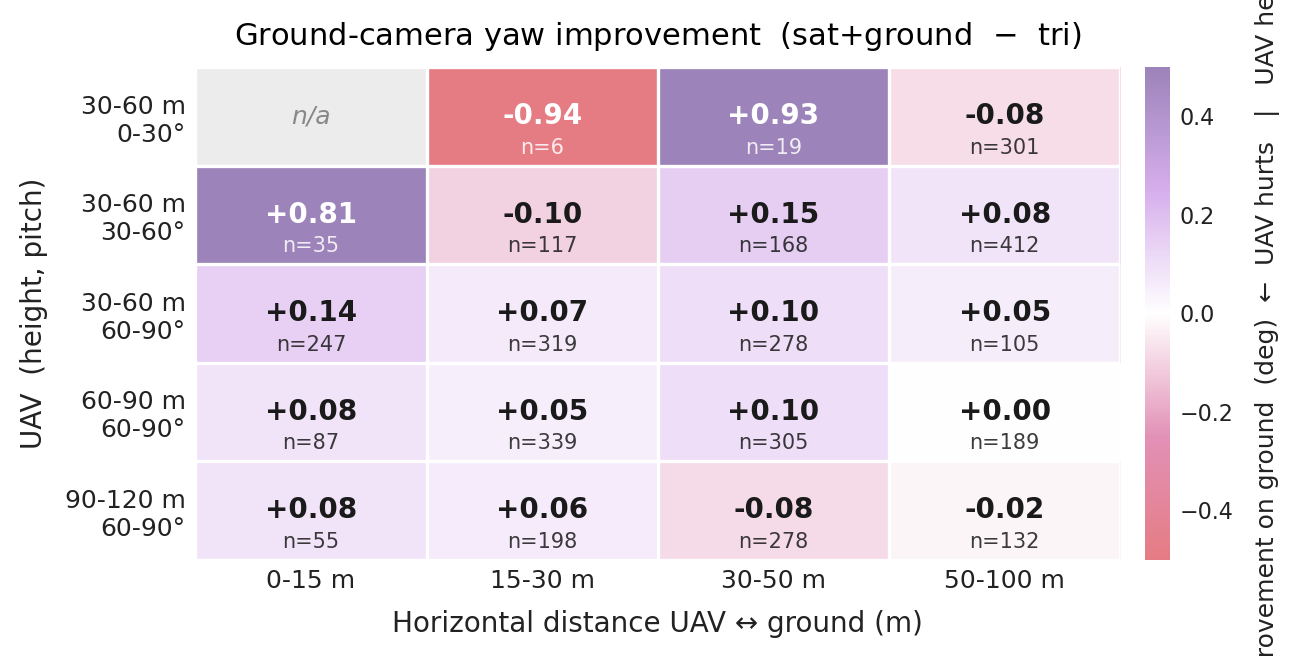}
  \caption{Per-cell value is $\text{metric}_{\text{sat+grd}}-\text{metric}_{\text{tri}}$ for ground position (\emph{left}, m) and ground yaw (\emph{right}, $^{\circ}$); red = the UAV helps, blue = the UAV hurts.}
  \label{fig:uav_heatmaps}
  \vspace{-4pt}
\end{figure}

\begin{figure}[!t]
\begin{minipage}[t]{0.49\textwidth}
  \centering
  \captionof{table}{Leave-one-out architectural and view-mode ablations on CrossGeo full test ($n{=}3{,}736$). Lower is better for every column.}
  \label{tab:arch_ablation}
  \vspace{1pt}
  \resizebox{\linewidth}{!}{%
  \renewcommand{\arraystretch}{0.95}
  \begin{tabular}{lccccc}
  \toprule
   & & \multicolumn{2}{c}{Ground} & \multicolumn{2}{c}{UAV} \\
  \cmidrule(lr){3-4} \cmidrule(lr){5-6}
  Method & Acc-mean$\downarrow$ & Met.Med$\downarrow$ & Yaw Med$\downarrow$ & Met.Med$\downarrow$ & Yaw Med$\downarrow$ \\
  \midrule
  \textbf{Cross3R}             & \textbf{1.07} & \textbf{2.14} & \textbf{1.30} & \textbf{1.64} & 0.87 \\
  w/o ortho                    & \underline{1.09} & 4.04 & 1.38 & 7.33 & \underline{0.77} \\
  w/o ms-fusion                & 1.16 & 2.24 & 1.35 & 1.79 & 0.82 \\
  w/o sat-pos-embed            & 1.16 & \underline{2.20} & \underline{1.35} & \underline{1.76} & \textbf{0.73} \\
  w/o sat-inject               & 1.12 & 2.36 & 1.61 & 1.83 & 0.78 \\
  w/o double register          & 1.12 & 2.43 & 1.39 & 1.80 & 0.82 \\
  w/o altitude-redef           & 3.89 & 7.37 & 1.48 & 12.00 & \underline{0.77} \\
  \midrule
  w/o UAV (sat$+$grd)          & 1.28 & 2.48 & 1.37 & --   & --   \\
  w/o ground (sat$+$UAV)       & 1.34 & --   & --   & 1.84 & 0.80 \\
  \bottomrule
  \end{tabular}}
\end{minipage}\hfill
\begin{minipage}[t]{0.49\textwidth}
  \centering
  \captionof{figure}{Cross3R data-scaling on CrossGeo (lowres $224\!\times\!224$, $n{=}3{,}736$).}
  \label{fig:data_scaling}
  \vspace{1pt}
  \includegraphics[width=\linewidth]{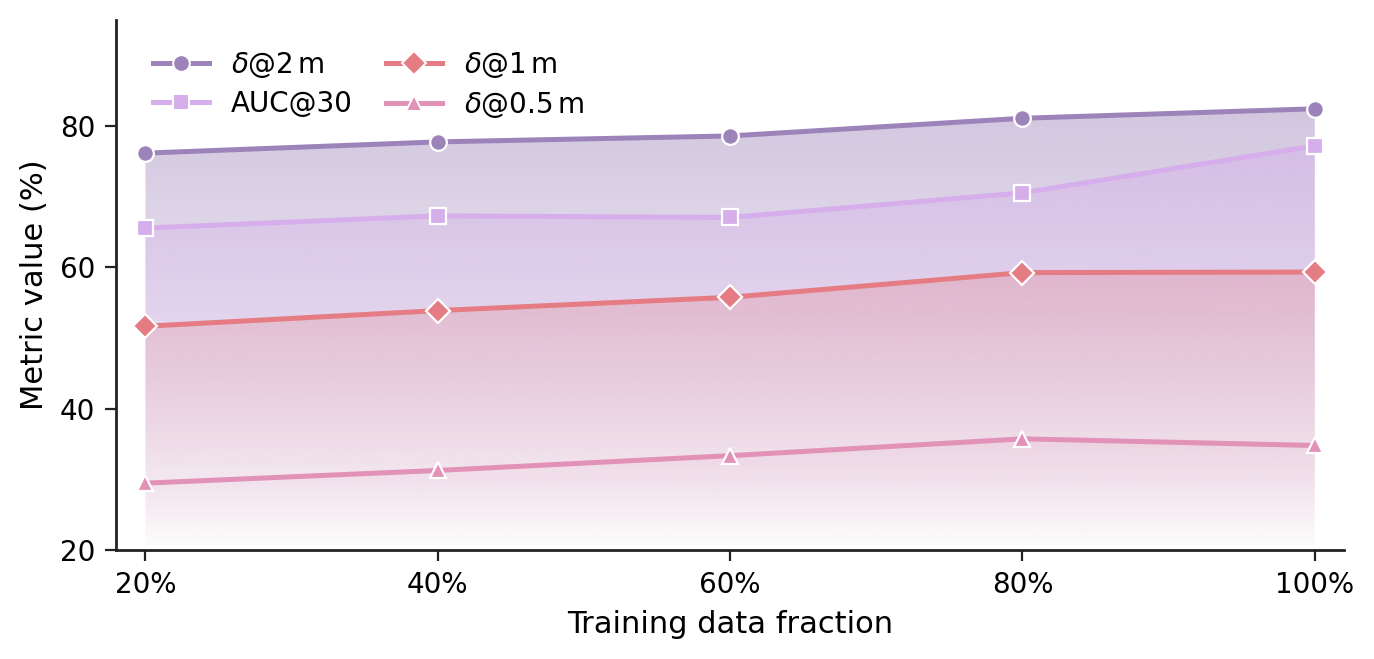}
\end{minipage}
\vspace{-12pt}
\end{figure}
\endgroup

\paragraph{Architectural components.} We isolate each design choice by training a leave-one-out variant of Cross3R on CrossGeo (Table~\ref{tab:arch_ablation}). \emph{Removing the orthographic head ($\rho$)} leaves Acc-mean nearly unchanged but causes cross-view localization to collapse---both UAV and ground translation errors increase sharply---confirming that $\rho$ is the essential link that ties the three branches into a shared, scale-consistent coordinate frame. \emph{Restoring the raw satellite altitude} ($\sim\!5{,}715$\,m, i.e., undoing the altitude redefinition) is catastrophic on every metric: the satellite branch must regress depths above $5{,}000$\,m while the UAV and ground branches operate within a few hundred meters, and the satellite camera sits two orders of magnitude higher than the perspective cameras---a depth and translation span the shared backbone cannot learn to bridge. The remaining four modules---multi-scale decoder fusion, Fourier positional embedding, satellite-to-perspective feature injection, and the doubled register-token bank---each bring a modest but consistent gain in point-cloud quality and ground/UAV translation; the injection module further improves ground-yaw accuracy. The bottom two rows of Table~\ref{tab:arch_ablation} evaluate the same checkpoint after dropping one perspective view at inference: excluding the UAV substantially degrades ground localization in both translation and yaw, whereas excluding the ground view only marginally affects UAV localization, mirroring the asymmetry seen in the heatmaps above (Fig.~\ref{fig:uav_heatmaps}) and consistent with the wider satellite--ground viewpoint gap.

\paragraph{Data scaling.} To validate the effectiveness of our dataset and assess the model's scaling potential, we train Cross3R at low resolution ($224\!\times\!224$) on $20\%$ - $100\%$ of the training data (Figure~\ref{fig:data_scaling}). Almost every metric improves monotonically, and the curves show no sign of saturating even at the full training set. In particular, from $80\%$ to $100\%$ data, the AUC@$30$ metric still exhibits a substantial gain. These trends indicate that our model has considerable headroom to benefit from additional data and confirm that our data acquisition pipeline is effective.

\section{Conclusion}
\label{sec:conclusion}
Cross3R exhibits strong cross-view reconstruction but several constraints remain. Due to limited training compute, each forward pass accepts at most three images. The network also handles only pinhole inputs, even though panoramic and fisheye cameras are increasingly vital for autonomous driving. Moreover, the model is data‑hungry, with clear headroom for improvement as the dataset scales. We plan to invest more compute to support more images and longer sequences, keep expanding CrossGeo with additional scenes, and broaden the network to diverse camera types. None of this changes our central message: through our acquisition pipeline and network design, we can recover satellite, drone, and ground point clouds, camera poses, and cross‑view localization in a single shot.
%%%%%%%%%%%%%%%%%%%%%%%%%%%%%%%%%%%%%%%%%%%%%%%%%%%%%%%%%%%%

\bibliographystyle{plainnat}
\bibliography{references}

\newpage
\appendix
\etocdepthtag.toc{supp}

% Supplementary table of contents (appendix only).
\renewcommand{\contentsname}{Supplementary Contents}
\begingroup
  \etocsettagdepth{main}{none}
  \etocsettagdepth{supp}{subsection}
  \tableofcontents
\endgroup
\bigskip

\section{Related Work}
\label{sec:Related_Work}

\paragraph{Cross-view Ground Image Localization.} This task estimates the pose of a ground-level camera within a candidate satellite region~\cite{shi2022beyond,xia2023convolutional,sarlin2023orienternet,sarlin2023snap,li2024bevformer,lee2025pidloc} and is challenging because of the large appearance and viewpoint gap between ground and aerial imagery. One line of work matches the two views in a learned feature space using contrastive objectives~\cite{tong2025geodistill,zhu2021vigor,wang2025bevsplat, shi2024weakly}, which is effective for coarse retrieval but struggles with fine-grained pose. A second line projects the ground view into a bird's-eye-view representation and establishes explicit geometric correspondences with the satellite image~\cite{song2023learning,wang2023fine,xia2025fg,wang2024view}, yielding more interpretable estimates. Both families share a fundamental limitation: a single nadir tile provides no cues for roll, pitch, or height, so they all adopt a planar-motion and zero-tilt assumption that reduces the problem to 3-DoF pose estimation and breaks on slopes, ramps, or tilted camera mounts. Cross3R lifts this restriction by adding a UAV image as an intermediate viewpoint that supplies the missing roll/pitch/altitude cues.

\paragraph{3D Scene Reconstruction.} Recovering 3D geometry and cameras from images is a long-standing problem. Classical Structure-from-Motion~\cite{schonberger2016structure, schoenberger2016vote, wang2024vggsfm, elflein2025light3r, zhao2025diffusionsfm, li2025cvd, pan2024global, lindenberger2021pixel} and Multi-View Stereo~\cite{furukawa2015multi, hiep2009towards, schops2017multi, yao2018mvsnet, duzceker2021deepvideomvs, brachmann2024acezero, izquierdo2025mvsanywhere, wang2026flying} pipelines, together with their modern learned variants, remain the workhorses for high-fidelity reconstruction, but require heavily overlapping views and degrade under sparse inputs or large viewpoint gaps. Recent feed-forward 3D models~\cite{wang2024dust3r, leroy2024grounding, yang2025fast3r, wang2025vggt, wang2025pi, keetha2025mapanything, wang2025continuous, liu2025slam3r, guo2026panovggt} drop per-scene optimization and instead predict geometry and cameras jointly in a single forward pass, which generalizes to novel scenes without per-instance tuning. Monocular depth foundation models~\cite{lin2025depth} supply reusable priors, and AerialMD~\cite{vuong2025aerialmegadepth} specializes DUSt3R to the aerial--ground setting by fine-tuning on a dedicated UAV-ground corpus. All of these assume image collections from a roughly common altitude; none handles a satellite/UAV/ground triplet---the gap Cross3R fills.

\paragraph{Multi-view Datasets.}
As summarized in Table~\ref{tab:datasets_comparison}, existing datasets cover only two modalities or omit poses. Ground-satellite benchmarks (CVUSA~\cite{workman2015localize}, CVACT~\cite{liu2019lending}, KITTI~\cite{geiger2013vision}, VIGOR~\cite{zhu2021vigor}) provide ground and satellite views but lack UAV images, restricting them to 3-DoF planar assumptions. UAV-focused datasets (SEUS-200~\cite{zhu2023sues}, GTA-UAV~\cite{ji2025game4loc}, UAV-VisLoc~\cite{xu2024uavvisloc}, AnyVisLoc~\cite{ye2025exploring}) include aerial imagery but omit ground-level views, missing the fine-grained street-level geometry. University-1652~\cite{zheng2020university} offers all three modalities but provides no camera poses, making it suitable only for image retrieval. Multi-view 3D datasets (ULTRRA~\cite{2zs6-ht63-24}, AerialMD~\cite{vuong2025aerialmegadepth}) offer 6-DoF poses but lack satellite imagery as a top-down reference. This absence has stalled joint satellite--UAV--ground reasoning, which CrossGeo closes with all three modalities, full 6-DoF poses, and dense metric depth.

\section{CrossGeo Dataset Details}
\label{sec:supp_dataset}

\subsection{Dataset Overview}
\label{sec:supp_dataset_overview}
\begin{figure}[!t]
  \centering
  \includegraphics[width=\textwidth]{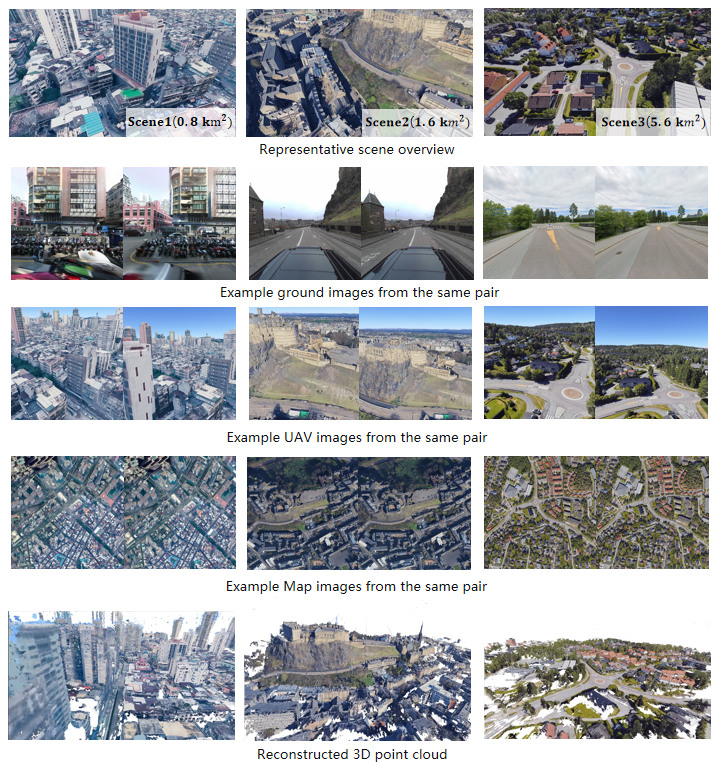}
  \caption{Overview of the five core modalities in the CrossGeo dataset. Row~1: representative scene overviews from three selected environments (Urban, Hilly, Rural), annotated with their coverage areas ($0.8$~km$^2$, $1.6$~km$^2$, $5.6$~km$^2$). Row~2: example ground images from the same triplet, showing dense buildings (Urban), undulating terrain and varied structures (Hilly), and flat open areas with low-rise constructions (Rural). Row~3: example UAV images from the same triplet, illustrating oblique aerial views across the three scene types. Row~4: example satellite map images (orthographic tiles) from the same triplet, providing top-down context. Row~5: reconstructed 3D point clouds for the corresponding scenes, visualizing the joint geometry recovered from satellite, UAV, and ground views.}
  \label{fig:dataset_summary}
\end{figure}

To provide an intuitive overview of the CrossGeo dataset, Figure~\ref{fig:dataset_summary} visualizes its five core modalities across three representative scene types: \textbf{Urban}, \textbf{Hilly}, and \textbf{Rural}. The Urban scene is characterized by dense high-rise buildings and bustling streets; the Hilly scene features undulating terrain with a mix of low-rise structures and natural vegetation; and the Rural scene consists of flat open land with sparse, low-rise constructions. This diversity in urban density, topography, and architectural styles directly supports the ``Multiple'' scenes claim for CrossGeo in Table~\ref{tab:datasets_comparison}.

\subsection{Dataset Coverage}
\label{sec:supp_coverage}

\begin{figure}[H]
  \centering
  \includegraphics[width=\textwidth]{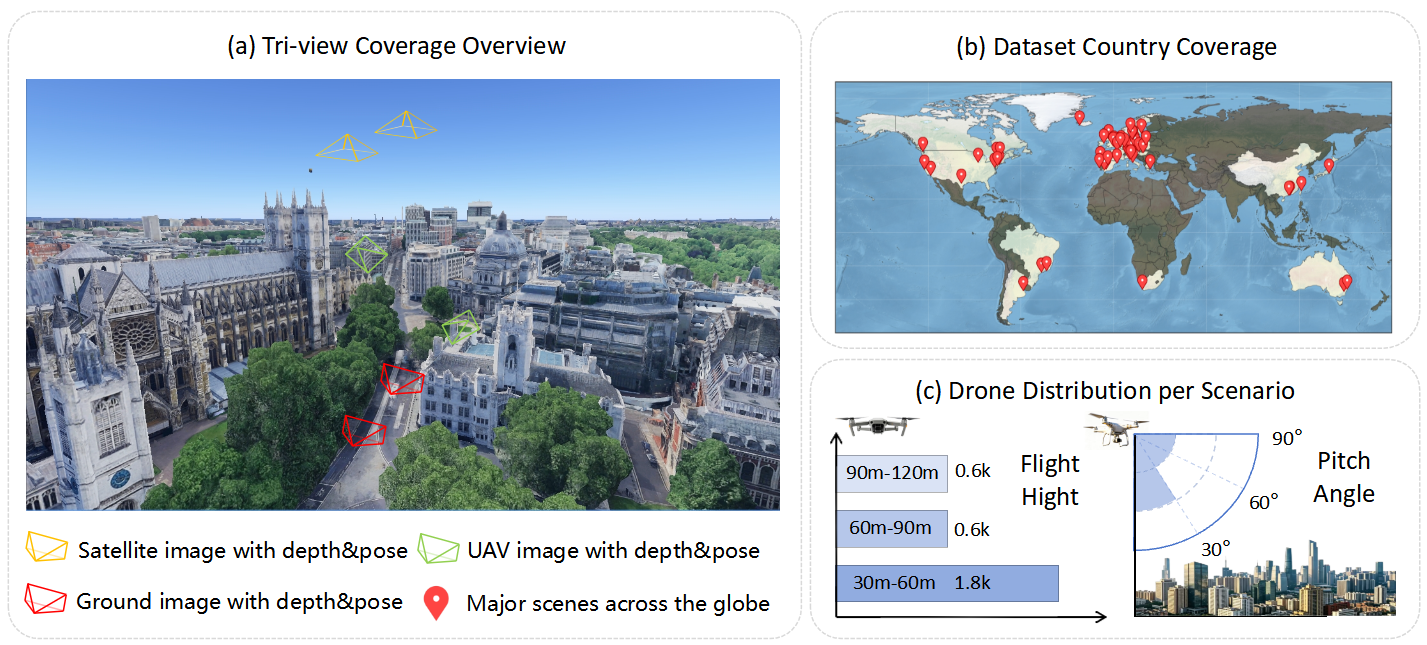}
  \caption{CrossGeo tri-view coverage and statistics: (a) a representative tri-view sample, (b) global geographic distribution of the 85 scenes, and (c) UAV altitude and pitch distributions.}
  \label{fig:dataset_overview}
\end{figure}

Figure~\ref{fig:dataset_overview} visualizes CrossGeo's tri-view coverage at the sample level along with its global geographic distribution and UAV-pose statistics, complementing the data-source overview in Figure~\ref{fig:data_sources} of the main paper. Panel~(a) shows a representative tri-view sample, in which the satellite (yellow), UAV (green), and ground (red) cameras are placed in a shared metric world frame, each carrying its own depth map and 6-DoF pose. Panel~(b) shows the geographic coverage: the red markers and highlighted countries mark the 85 scenes, spanning every continent except Antarctica and a wide mix of urban, suburban, rural, and natural terrain. Panel~(c) summarizes the UAV pose distribution per scenario---flight altitudes are dominated by the $30$--$60$\,m bin (about $1.8$\,k samples), with smaller bins at $60$--$90$\,m and $90$--$120$\,m (about $0.6$\,k each); pitch angles span $0^{\circ}$--$90^{\circ}$, with the low-pitch range ($0^{\circ}$--$30^{\circ}$) oversampled to mitigate the pixel stretching that high-altitude high-pitch renderings produce.

\subsection{Street-View Image Release Policy}
\label{sec:supp_release}

The street-level ground images in CrossGeo and AnyVisLoc are pulled from Google Maps\footnote{\url{https://www.google.com.hk/earth/education/tools/street-view/}} and Baidu Maps\footnote{\url{https://lbsyun.baidu.com/products/map}} through their official APIs. Because of privacy protection and the providers' terms of service, we are not allowed to redistribute these street images ourselves. Instead, for every ground sample we release the image's \texttt{pano\_id} (the unique ID that Google and Baidu assign to each street image), along with the camera pose and capture metadata we used. With this \texttt{pano\_id}, anyone who has API access can call the same Google Maps or Baidu Maps endpoint, follow each provider's usage rules, and download exactly the same street image we trained and evaluated on. The UAV captures, satellite tiles, and all derived depth maps and poses are released directly as part of CrossGeo.

\newpage
\subsection{Depth Acquisition Pipeline}
\label{sec:supp_pipeline}

\begin{figure}[!t]
  \centering
  \includegraphics[width=\textwidth]{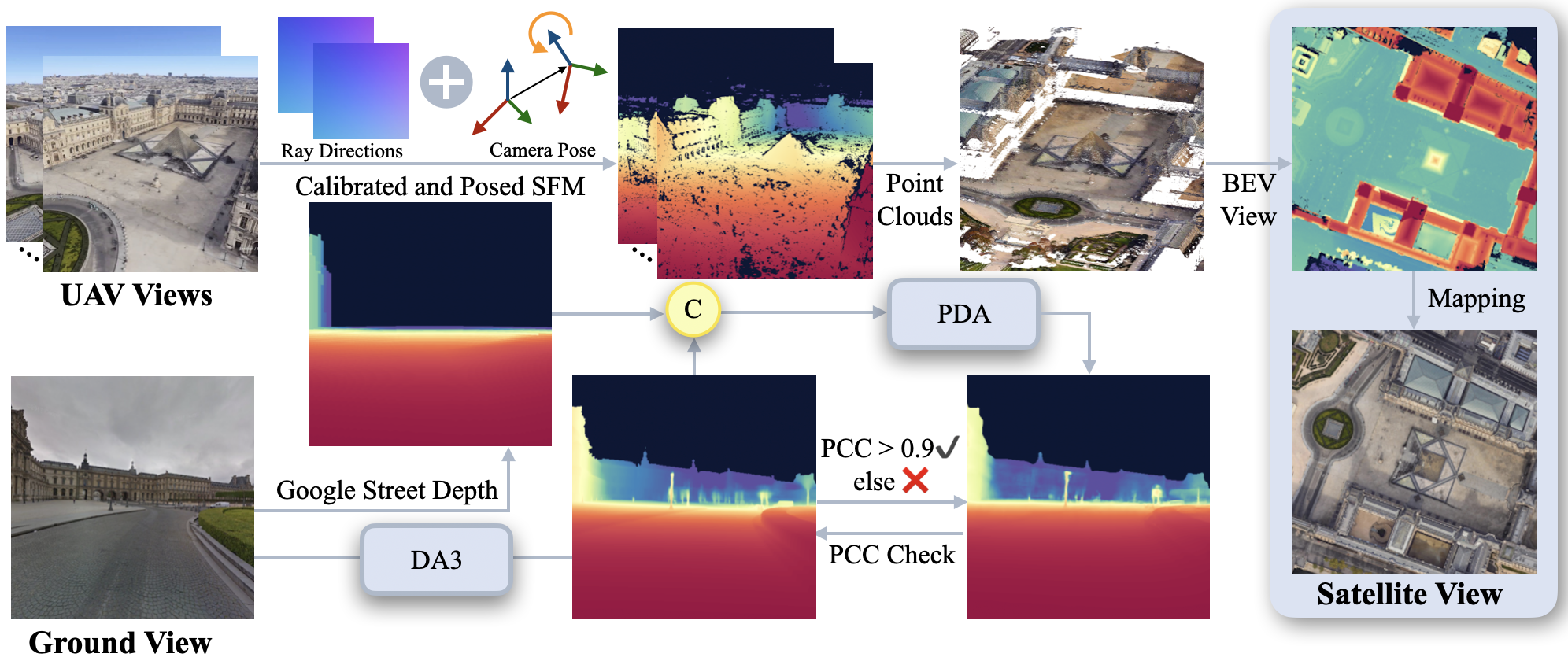}
  \caption{CrossGeo depth-acquisition and tri-view alignment pipeline: each modality produces a dense metric depth map and is registered into a shared coordinate frame.}
  \label{fig:data_pipeline}
\end{figure}

Figure~\ref{fig:data_pipeline} details the full processing pipeline we use to recover dense metric depth for every modality and to register the three views into a shared coordinate frame. \emph{UAV branch.} Captures are processed by a calibrated, posed Structure-from-Motion stage~\cite{schonberger2016structure} with known ray directions and camera poses, producing a dense metric UAV point cloud whose BEV projection yields per-pixel UAV depth. \emph{Ground branch.} Each view is paired with the coarse, absolute-scale depth from \emph{Google Street View} and refined by a monocular depth backbone (DA3)~\cite{lin2025depth}; the two depth signals are concatenated (\textbf{C}) and fused into a metric ground depth by a pose--depth alignment module (\textbf{PDA})~\cite{wang2025depth} that uses the Street View depth as a metric anchor for the high-resolution but relative DA3 prediction. \emph{Satellite branch.} Tiles are sourced from \emph{Google Maps}; we associate them with the UAV point cloud's BEV depth to obtain a metric depth for each satellite pixel. The output is paired (satellite, UAV, ground) views with consistent poses and dense metric depth across all three modalities.

\begin{figure}[!t]
  \centering
  \includegraphics[width=\textwidth]{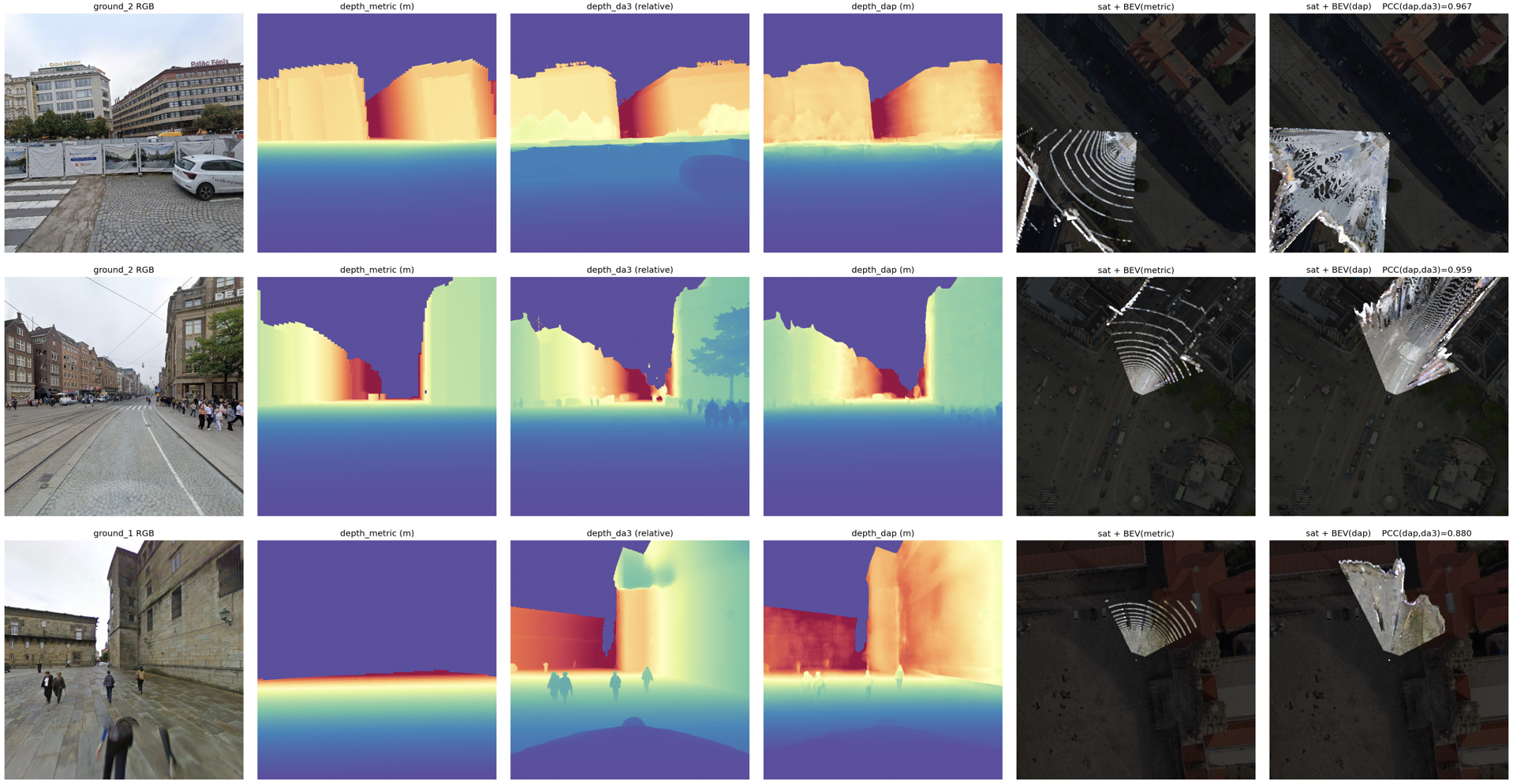}
  \caption{Qualitative ground-depth samples and the PCC quality filter. \emph{Top two rows}: successful fusions (PCC$>$$0.9$) that are kept. \emph{Bottom row}: a failure case (PCC$<$$0.9$) that is discarded, caused by missing building depth in the Google Street View anchor.}
  \label{fig:depth_sample}
\end{figure}

Figure~\ref{fig:depth_sample} shows representative qualitative results of the ground-depth fusion together with our PCC-based quality filter. The top two rows are successful samples whose rescaled depth has a Pearson correlation~\cite{benesty2009pearson} greater than $0.9$ with the relative DA3~\cite{lin2025depth} prediction; the geometry of buildings, road surface, and street furniture is faithfully recovered and the sample is kept. The bottom row is a failure case with PCC below $0.9$ and is therefore discarded: the Google Street View depth on which we anchor the metric scale lacks any depth values for the buildings, so the fused ground depth collapses to an incorrect plane that disagrees with the relative DA3 prediction in regions where buildings dominate the view.

\section{Satellite Altitude Ablation}
\label{sec:supp_sat_altitude}

Training is unstable at the raw $5{,}726$\,m tile altitude (Tab.~\ref{tab:arch_ablation}), so before training we have to pick a redefined satellite altitude. We try three values---$100$, $150$, and $200$\,m---and retrain Cross3R end-to-end at each setting, keeping everything else the same and evaluating on the same CrossGeo test set (Table~\ref{tab:sat_altitude_ablation}). Point-cloud quality stays the same across all three runs, so this choice does not really affect the 3D geometry. Camera pose, on the other hand, is very sensitive to the altitude. At $100$\,m the satellite virtual camera ends up inside the UAV altitude range ($30$--$120$\,m), so the satellite no longer offers a clear top-down view that is different from the UAV; the model loses its top-down anchor and translation accuracy drops sharply. At $200$\,m the satellite sits far above the UAV and ground views, and the gap between the satellite depth range and the perspective depth range becomes large enough that the model struggles to match them in scale. $150$\,m sits between these two cases: it is high enough to stay above every UAV altitude in our data, and low enough that the satellite, UAV, and ground branches still share a similar depth range---the condition Tab.~\ref{tab:arch_ablation} already showed to be necessary. We therefore use $150$\,m as the redefined satellite altitude in every main-paper experiment.

\begin{table}[!t]
\centering
\caption{Satellite-altitude ablation on CrossGeo (lowres $224\!\times\!224$, $n{=}3{,}736$). All rows share the Cross3R architecture and the same compute budget; only the redefined satellite altitude changes. Best in \textbf{bold}.}
\label{tab:sat_altitude_ablation}
\setlength{\tabcolsep}{4pt}
\renewcommand{\arraystretch}{1.05}
\resizebox{\textwidth}{!}{%
\begin{tabular}{lcccc|ccccc}
\toprule
 & \multicolumn{4}{c|}{Point cloud} & \multicolumn{5}{c}{Camera pose (\%)} \\
\cmidrule(lr){2-5} \cmidrule(lr){6-10}
Altitude & Acc-mean$\downarrow$ & $\delta@0.5$m$\uparrow$ & $\delta@1$m$\uparrow$ & $\delta@2$m$\uparrow$ & RRA@$5^{\circ}$ & RRA@$15^{\circ}$ & RTA@$5^{\circ}$ & RTA@$15^{\circ}$ & AUC@$30$ \\
\midrule
$100$\,m                       & \textbf{1.27} & 35.06 & 59.29 & 82.35 & \textbf{84.85} & \textbf{93.92} & 1.39  & 52.81 & 44.10 \\
$150$\,m \textbf{(Cross3R)}    & \textbf{1.27} & 35.25 & \textbf{59.93} & 82.21 & 84.64 & 92.37 & \textbf{70.85} & \textbf{97.11} & \textbf{79.85} \\
$200$\,m                       & \textbf{1.27} & 34.79 & 59.32 & \textbf{82.41} & 83.89 & 92.16 & 60.25 & 95.48 & 77.18 \\
\bottomrule
\end{tabular}}
\end{table}

\section{Failure Cases on CrossGeo}
\label{sec:supp_failure_cases}

\begin{figure}[H]
  \centering
  \includegraphics[width=\textwidth]{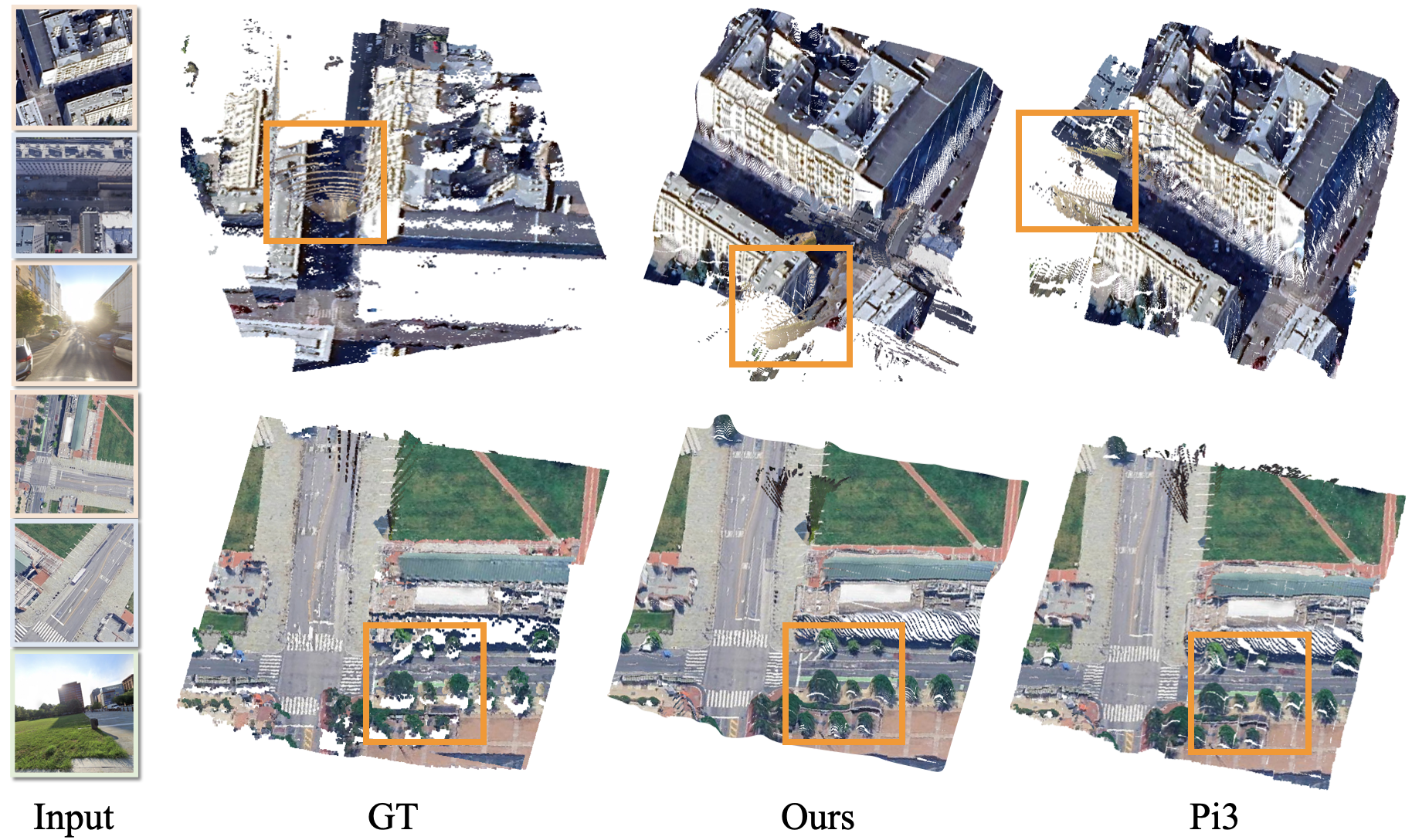}
  \caption{Failure (\emph{top row}) and success (\emph{bottom row}) cases of Cross3R on CrossGeo.}
  \label{fig:more_case}
\end{figure}

Figure~\ref{fig:more_case} contrasts a failure case (top row) with a successful case (bottom row) drawn from the CrossGeo test set. \emph{Top row.} When the ground image lacks clear localization cues---for example, a long, straight street with little distinctive structure and poor lighting that washes out the road surface---Cross3R can be drawn to a visually similar but geometrically wrong location on the satellite tile. This is a well-known and largely unavoidable failure mode of any cross-view localization method, since under such conditions the ground image is genuinely ambiguous with respect to the satellite tile. \emph{Bottom row.} When the ground view is well-lit and contains discriminative structures (curbs, building facades, road markings), Cross3R produces both an accurate ground-camera pose and a high-quality reconstruction: the ground point cloud aligns almost seamlessly with the satellite tile, illustrating that the model can fully exploit reliable visual cues when they are available.

\section{Qualitative Results on KITTI}
\label{sec:supp_qual_kitti}

\begin{figure}[H]
  \centering
  \includegraphics[width=\textwidth]{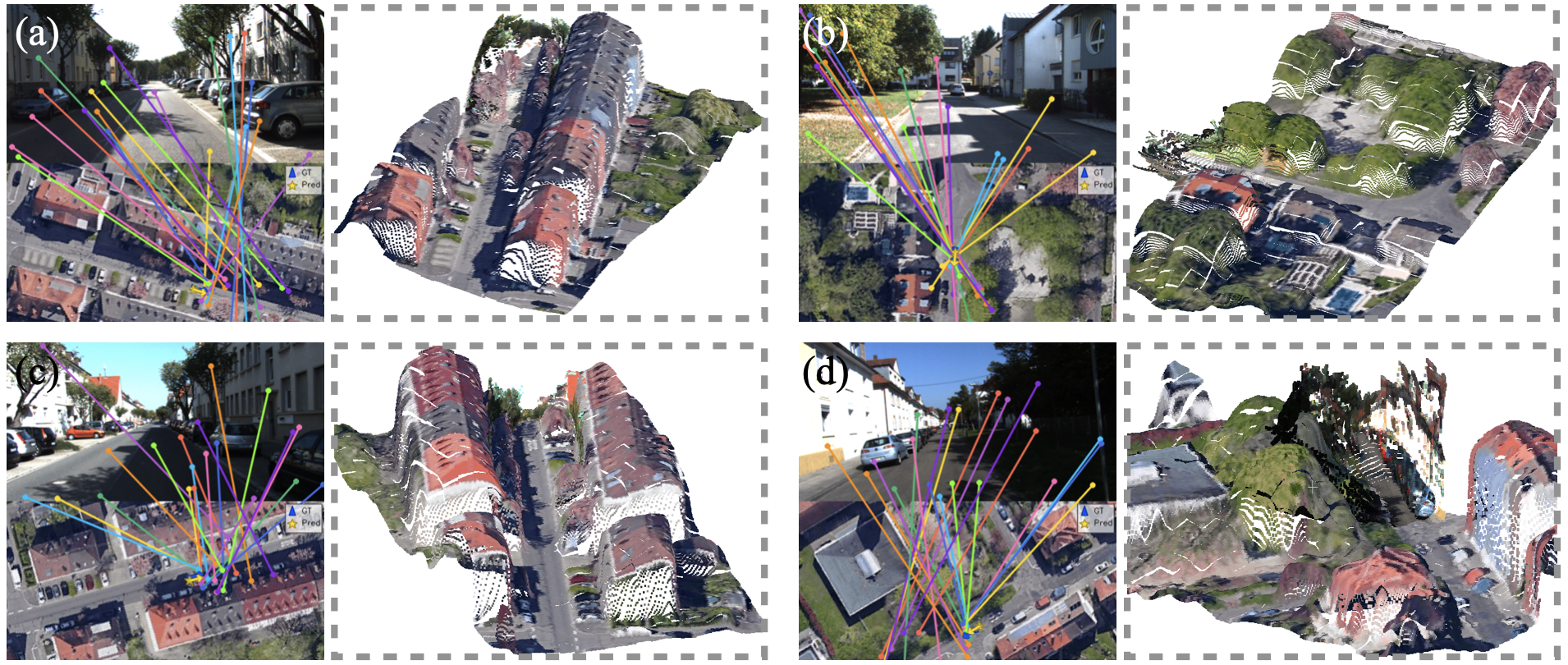}
  \caption{Zero-shot cross-view localization, pixel matching, and tri-view reconstruction on KITTI, samples (a)--(d). \emph{Left of each panel}: ground-to-satellite localization and pixel-matching. \emph{Right of each panel}: corresponding stacked point cloud.}
  \label{fig:kitti_vis}
\end{figure}

Figure~\ref{fig:kitti_vis} visualizes four KITTI~\cite{geiger2013vision} samples (a)--(d): each sample's left half shows ground-to-satellite localization and pixel matching, the right half the stacked point cloud. The ground reconstruction lands cleanly on top of the satellite branch, and is nearly seamless in (a) and (c). All predictions are zero-shot---Cross3R is never trained on KITTI---corroborating Tab.~\ref{tab:kitti_crossarea} of the main paper.

\section{Qualitative Results on AnyVisLoc}
\label{sec:supp_qual_anyvisloc}

\begin{figure}[H]
  \centering
  \includegraphics[width=\textwidth]{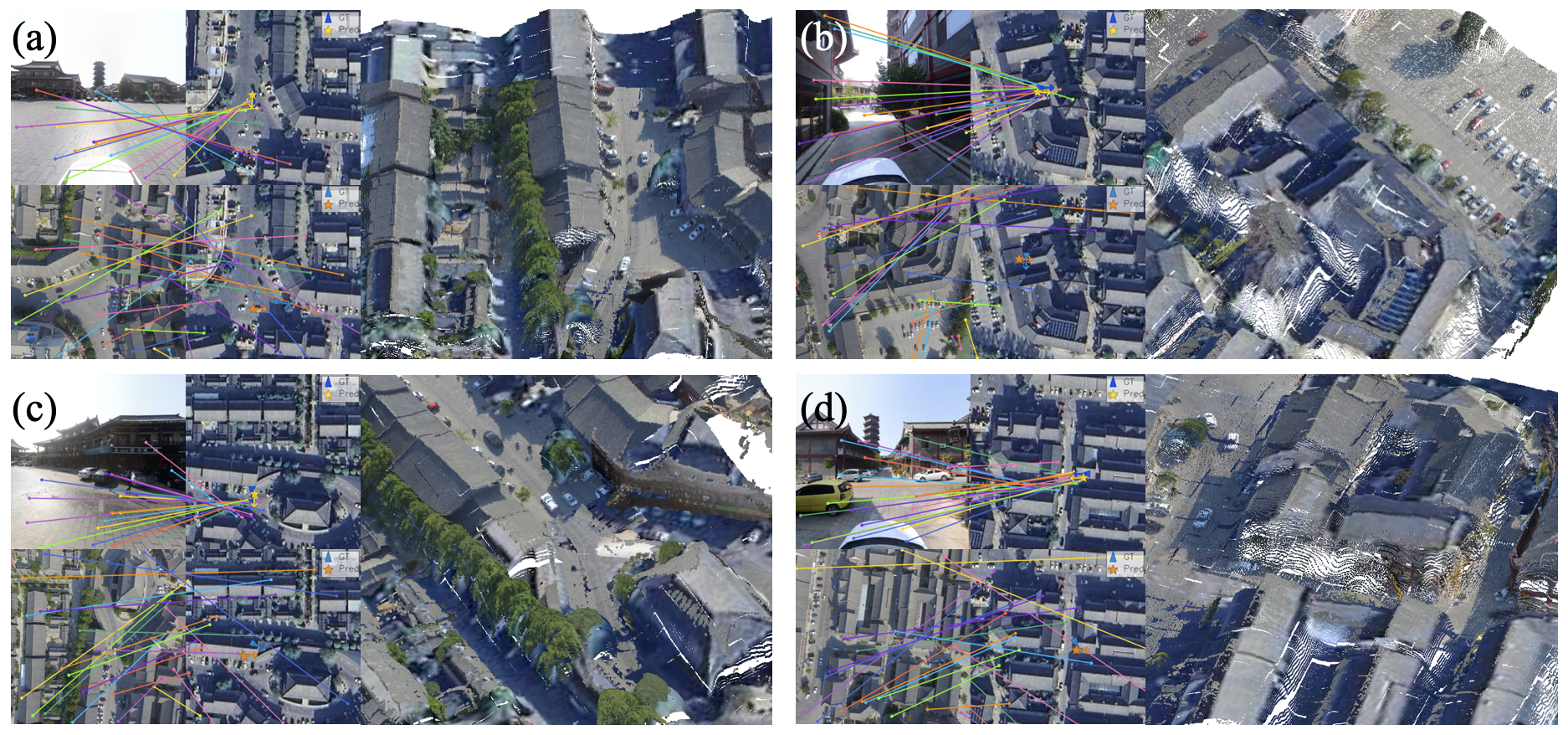}
  \caption{Out-of-distribution cross-view localization, pixel matching, and tri-view reconstruction on AnyVisLoc. For each sample: \emph{top-left}: ground-to-satellite, \emph{bottom-left}: UAV-to-satellite, \emph{right}: stacked tri-view point cloud.}
  \label{fig:anyvisloc_vis}
\end{figure}

Figure~\ref{fig:anyvisloc_vis} visualizes Cross3R on four AnyVisLoc~\cite{ye2025exploring} samples; for each sample the top-left panel shows ground-to-satellite localization and pixel matching, the bottom-left panel the corresponding UAV-to-satellite result, and the right half the predicted tri-view point cloud. Although Cross3R wins most quantitative columns on AnyVisLoc (Tab.~\ref{tab:sota_anyvisloc}), the qualitative point clouds are noticeably noisier than on CrossGeo and KITTI for two reasons specific to this benchmark: the Google Maps tiles in these regions are served at a lower resolution and the buildings are visually repetitive, so the satellite branch carries less localization signal; and most AnyVisLoc UAV photographs are captured above $120$\,m, outside CrossGeo's $30$--$120$\,m training range. Cross3R nonetheless produces a coherent tri-view layout, which we consider acceptable for a feed-forward model evaluated entirely out of distribution.

\section{Broader Impact}
\label{sec:broader_impact}

Cross3R aims to push cross-view ground-camera localization beyond the 3-DoF limit of prior methods, enabling reliable operation in real-world settings where the ground camera may be tilted or placed on uneven terrain. We envision applications such as autonomous driving on rough roads, robot navigation in unfamiliar environments, and search-and-rescue missions where a ground team collaborates with a drone. At the same time, accurate cross-view localization can be misused: a single ground photo could be localized without consent, enabling unwanted tracking or surveillance. To mitigate this risk, we built CrossGeo without any personally identifiable annotations. We also do not redistribute the original Google Street View images; we release only the \texttt{pano\_id} for each image, so that anyone using the data must fetch it through Google's official API and comply with Google's terms of service. We urge those building upon Cross3R to respect the privacy and consent policies of the data providers and the legal requirements of the regions where the model is deployed.

%%%%%%%%%%%%%%%%%%%%%%%%%%%%%%%%%%%%%%%%%%%%%%%%%%%%%%%%%%%%

% NeurIPS checklist is omitted from the arXiv preprint version.
% \newpage
% \input{checklist.tex}

\end{document}